%% file: main.tex
\ifdefined\pdfinclusioncopyfonts
\fi
\documentclass{prismshadow}
\usepackage[T1]{fontenc}
\input{package}

\title{From Existing Methods to\\New Research Ideas: Two Perspectives\\on Recursive Self-Improvement}
\author[1,*]{Leijun~Zhou}
\author[1,*]{Yanke~Yu}
\author[1]{Zhihao~Liu}
\author[1]{Qu~Xiang}
\author[2]{Tiancheng~Hu}
\author[1]{Yiru~Dong}
\author[1]{Yihan~Wu}
\author[3]{Yangxi~Xiang}
\author[1]{Xuejun~Wu}
\author[4]{Xi~Chen}
\author[1]{Buyue~Qian}
\author[1]{Yaowei~Zheng}
\author[2]{Junhao~Hu}
\affiliation[1]{PrismShadow}
\affiliation[2]{Peking~University}
\affiliation[3]{Zhejiang~University}
\affiliation[4]{New~York~University}
\contribution[*]{Equal contribution.}
\abstract{\input{sections/en/abstract.tex}}
\date{October 9, 2026}
\correspondence{\mbox{Junhao~Hu (\email{hu@prismshadow.com})}, \mbox{Yaowei~Zheng (\email{zheng@prismshadow.com})}}

\begin{document}
\maketitle
\clearpage
\begingroup
\small
\renewcommand{\baselinestretch}{1}\selectfont
\setlength{\parskip}{0pt}
\tableofcontents
\endgroup
\clearpage
\input{sections/en/intro.tex}
\input{sections/en/background.tex}
\input{sections/en/framework.tex}
\input{sections/en/connections.tex}
\input{sections/en/evaluation.tex}
\input{sections/en/conclusion.tex}
\clearpage
\bibliographystyle{unsrtnat}
\bibliography{references}
\clearpage
\appendix
\input{sections/en/catalogue.tex}
\end{document}

%% file: package.tex
\usepackage{graphicx,amsmath,amssymb,booktabs,longtable,array}
\usepackage{bookmark,enumitem,needspace,pdflscape}
\setcitestyle{numbers,square,comma}
\setlist{nosep,leftmargin=*}

%% file: sections/en/abstract.tex
Recursive self-improvement (RSI) offers a route to AI systems that participate in their own continued development. Recent methods allow language agents to retain changes to their parameters, memory, skills, and code for subsequent execution and improvement. We examine how these methods can inform new research ideas through two complementary perspectives. First, we introduce a unified taxonomy with nine design dimensions organized into four groups: update targets, update organization, update schedule and scope, feedback and selection. This taxonomy places existing methods in a shared design space, where examining individual dimensions and their interactions can inspire new research ideas. Second, we examine how existing RSI methods draw on traditional deep learning, for example through optimization, regularization, data augmentation, continual learning, and adversarial learning. We develop further ideas by examining more specific learning mechanisms and drawing on evolutionary computation. We also review how existing benchmarks assess self-improvement under online, offline, and offline-to-online settings. Together, these perspectives help researchers develop new RSI research ideas from existing designs and learning mechanisms. RSI papers and resources are available at {\hypersetup{urlcolor=black}\url{https://prism-shadow.github.io/awesome-rsi/}}.

%% file: sections/en/intro.tex
\section{Introduction}
\label{introduction}

\begin{figure}[!htbp]
\centering
\includegraphics[width=\linewidth]{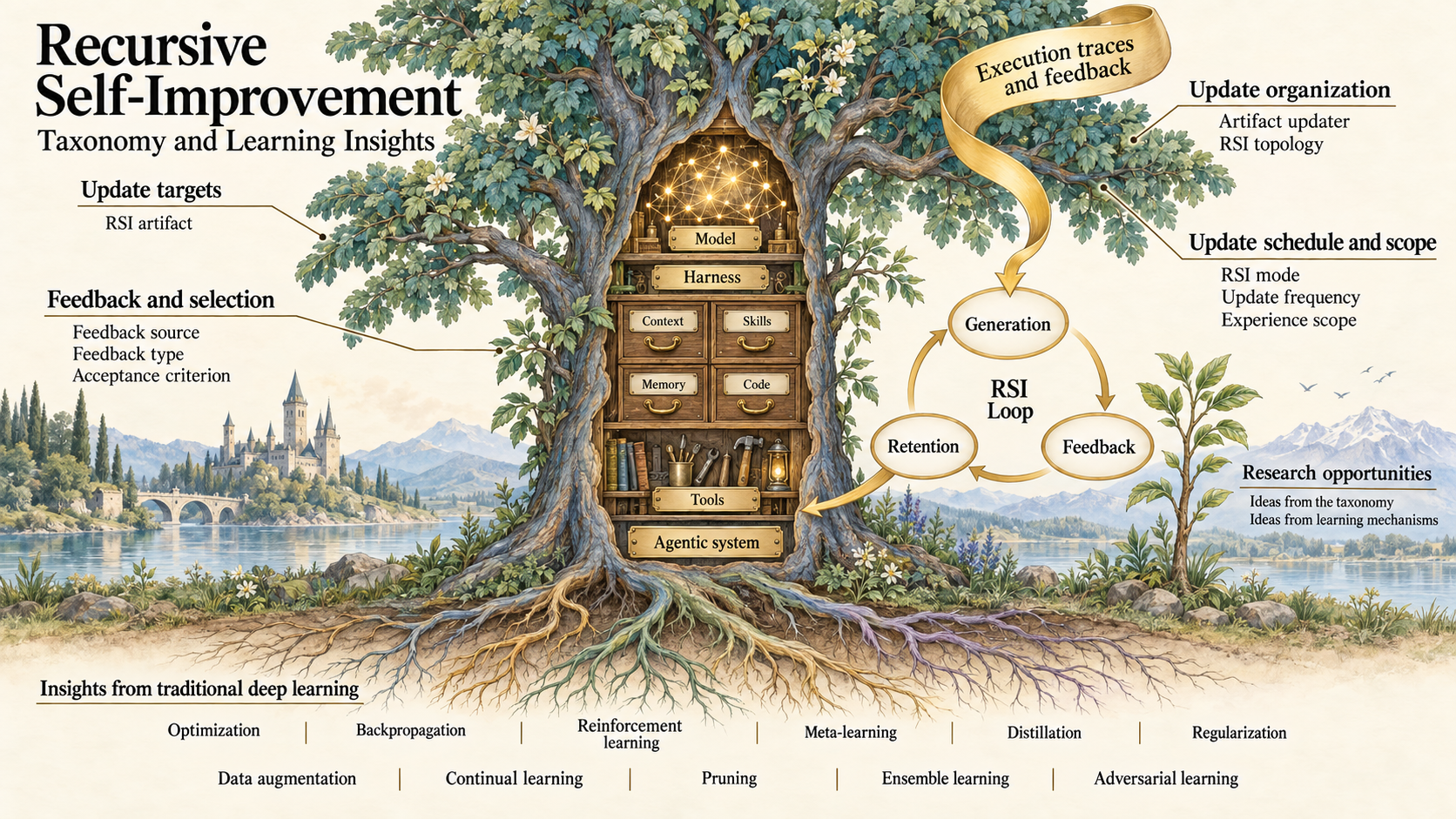}
\caption{Overview of the survey. Agent resources form the trunk, four groups of design dimensions organize the canopy, and traditional deep learning supplies the roots. Execution traces and feedback inform self-improvement. New growth represents research opportunities arising from the taxonomy and learning mechanisms.}
\label{fig:overview}
\end{figure}

Recursive self-improvement (RSI) is a process in which an AI system participates in modifying itself, retains these modifications, and uses the revised system in subsequent execution and improvement. Its appeal lies in the possibility that an improvement makes the system better at producing further improvements. A growing share of AI development could then be carried out by the systems being developed. This possibility underlies Good's suggestion that a sufficiently capable machine might be the last invention humans need to make \citep{good1965}.

Recent research has produced numerous RSI methods, including those that update model parameters, revise memory and skills, and modify agent code \citep{r2506_10943,memrl,r2609_08944,r2505_22954}. Several surveys now organize this literature by improvement targets, autonomy levels, or the degree of loop closure \citep{duan2026,chen2026,lisurvey}. These classifications help describe the field, but assigning a method to a broad category does not fully explain its design. Two methods that revise agent code, for example, can differ in who proposes changes, how earlier versions are reused, and what feedback determines which changes are retained. Generating new research ideas from existing work requires examining its design dimensions and their interactions to identify limitations of current mechanisms and new ways to improve them. Existing surveys also discuss connections to earlier learning paradigms, but offer less systematic guidance on using specific mechanisms from traditional deep learning to generate new RSI research ideas. This calls for an analysis of how RSI adapts these mechanisms and which principles can be reused.

We view existing RSI methods as combinations of design choices. These choices define a high-dimensional design space: each dimension captures an aspect of the improvement process, and each method occupies a point determined by its choices along these dimensions. Locating existing work in this space allows us to examine individual dimensions and their interactions, identify limitations of current mechanisms, and develop new research ideas for agent improvement.
To characterize this space, we introduce a unified taxonomy with four groups of design dimensions: \emph{update targets}, \emph{update organization}, \emph{update schedule and scope}, and \emph{feedback and selection}. These groups encompass nine dimensions, which we use to describe representative methods, allowing methods developed for different settings to be compared within the same space. The accompanying Awesome RSI website lets readers browse and filter methods along these dimensions.\footnote{GitHub repository: {\hypersetup{urlcolor=black}\url{https://github.com/Prism-Shadow/awesome-rsi}}.}

Traditional deep learning provides a second source of research ideas for RSI. It has developed mechanisms for choosing updates, learning from feedback, and controlling overfitting, all of which are relevant to recursive self-improvement. Recent work already draws on these connections: SkillAdam adapts ideas from Adam to skill revision; TextGrad applies the idea of backpropagation to textual feedback; and RRSI regularizes changes to agent code \citep{r2609_08944,textgrad,rrsi}. We study these correspondences to understand which principles carry over to RSI. We then examine more specific mechanisms within these connections and draw on evolutionary computation to develop further research ideas.

We also review benchmarks and evaluation settings for online, offline,
and offline-to-online improvement. This accompanying analysis explains
what each setting measures and motivates shared evaluation protocols.

Our contributions are twofold.
\begin{itemize}
\item \textbf{A unified taxonomy of agent self-improvement.}
We introduce nine design dimensions organized into four groups, grounded
in a formal description of the self-improvement loop. This taxonomy
places existing methods in a common design space, allowing us to compare
their mechanisms and examine relationships across dimensions. Through
a concrete example, we illustrate how this analysis can reveal
limitations of existing designs and inspire new RSI research ideas.

\item \textbf{Insights from traditional deep learning.}
We analyze how existing RSI methods draw on established learning
mechanisms, for example through optimization, regularization, data
augmentation, continual learning, and adversarial learning. We explain
the problems these mechanisms originally address and how RSI methods
adapt them to self-improvement. Building on this analysis, we explore
how deeper developments of these mechanisms, together with ideas from
broader machine learning and evolutionary algorithms, can generate
further RSI research ideas.
\end{itemize}

%% file: sections/en/background.tex
\section{Background}
\label{background}

We consider an agentic system to perform recursive self-improvement (RSI)
when it modifies itself, retains the changes, and uses the updated system
to execute subsequent tasks and produce further self-modifications. We
first introduce the agentic system and its task-execution loop, then
formalize the self-improvement loop and connect it to the design
dimensions developed in our taxonomy.

\subsection{Agentic systems}
\label{sec:agentic-systems}
\label{sec:background-organization}

\begingroup
\clubpenalty=10000
\clubpenalties 1 10000
An agentic system consists of a language model and a harness. The harness
comprises tools, context, memory, skills, and executable code. Tools are
callable operations through
which the system obtains information and acts on the environment. Context
is the information assembled for a model call, including instructions,
task inputs, and relevant interaction history. Memory stores information
beyond the current context for retrieval and reuse. Skills package
task-specific instructions with optional scripts and reference material
as reusable resources. Executable code coordinates model calls, context
construction, tool execution, and the use of memory and skills. The
environment refers to the external systems that the agent can observe
and affect.
\par\endgroup

\subsection{The ReAct Loop}
\label{chatbot}
\label{agent}

A basic chatbot produces an answer from a query, the supplied context,
and knowledge stored in its parameters. An agent extends this setting
by taking actions and using the resulting observations to decide what
to do next. It can obtain information, execute operations, and revise its
plan as the task proceeds. Feedback can also guide revision of an
attempted solution \citep{selfrefine}. ReAct organizes this interaction through the
interleaving of reasoning and acting \citep{react}. We use \emph{ReAct Loop}
to describe this task-execution cycle.

In each iteration, the harness assembles the context for a model call and
executes the tool operations requested by the model. Observations from
the environment become part of the information available for subsequent
decisions. Repeated interaction produces execution traces and task outcomes
that can inform changes to the system itself.

\subsection{The RSI Loop}
\label{recursive-self-improvement}

The \emph{RSI Loop} connects self-modification to subsequent execution
and improvement. The system gathers information, constructs candidate
changes, and incorporates selected changes into the configuration used
in later rounds. Task execution supplies information for this process.

We formalize the loop to describe how the current system gives rise to
the system used in the next round. Let $\mathcal A_t$ denote the current
system configuration. The update procedure $U$ covers a complete round of
self-improvement:
\begin{equation}
U\bigl(\mathcal A_t;D\bigr)
\longrightarrow\mathcal A_{t+1},\qquad t=0,1,2,\ldots.
\label{eq:rsi-update}
\end{equation}
Here, $t$ counts improvement rounds, and $D$ collects the method's nine
design settings. Within $U$, the functions $P$, $F$, and $R$ perform
candidate generation, candidate feedback, and retention in sequence.
Their connection can be written as
\begin{equation}
U(\mathcal A_t;D)
=R\!\left(\mathcal A_t,(F\circ P)(\mathcal A_t)\right).
\label{eq:rsi-composition}
\end{equation}
The settings of $P$, $F$, and $R$ are omitted in this overview and
specified below. The composition $F\circ P$ generates a candidate and
returns it together with its feedback. $R$ uses this pair and the current
system to determine the system retained for the next round.

\textbf{System configuration.}
Following the components described in Section~\ref{sec:agentic-systems},
we write
\begin{equation}
\mathcal A_t=(\mathcal M_t,\mathcal H_t),
\label{eq:rsi-system}
\end{equation}
where $\mathcal M_t$ contains the system's language models and their
parameters. The harness resources are
\begin{equation}
\mathcal H_t=\bigl(\mathrm{Tools}_t,\mathrm{Context}_t,
\mathrm{Memory}_t,\mathrm{Skills}_t,\mathrm{Code}_t\bigr).
\label{eq:rsi-harness}
\end{equation}
The entries correspond to the tools, context, memory, skills, and
executable code used in round $t$.

\textbf{Update schedule and scope.}
The first three entries of $D$ specify when $U$ operates and where its
retained changes are intended to be reused:
\begin{equation}
D=\bigl(d_{\mathrm{mode}},d_{\mathrm{frequency}},
d_{\mathrm{scope}},\ldots\bigr).
\label{eq:rsi-schedule-scope}
\end{equation}
\emph{RSI mode}, $d_{\mathrm{mode}}$, places improvement in an offline
phase, during online use, or first offline and then online.
\emph{Update frequency}, $d_{\mathrm{frequency}}$, specifies the amount
of execution or the event that triggers an update to persistent
resources: a step, a specified event, a complete trajectory, or a batch.
For example, a system can receive feedback after every task and revise
its skills after a batch of tasks. \emph{Experience scope},
$d_{\mathrm{scope}}$, describes whether retained resources are intended
for a particular task context or for reuse across task types and domains.
The remaining entries of $D$ describe the three functions within $U$.

\textbf{Candidate generation.}
$P$ uses the current system's resources together with available task
material, execution records, and feedback to construct a candidate
configuration $\widetilde{\mathcal A}_t$:
\begin{equation}
P\bigl(\mathcal A_t;
d_{\mathrm{artifact}},d_{\mathrm{updater}},d_{\mathrm{topology}}\bigr)
\longrightarrow\widetilde{\mathcal A}_t.
\label{eq:rsi-proposal}
\end{equation}
Three further entries of $D$ describe candidate generation.
\emph{RSI artifact}, $d_{\mathrm{artifact}}$, specifies which resources
are modified. \emph{Artifact updater}, $d_{\mathrm{updater}}$, identifies
the roles that form the update, including those that propose, revise,
or curate candidate changes. \emph{RSI topology},
$d_{\mathrm{topology}}$, describes how the candidate draws on earlier
versions and their relationships.

Candidates can revise memory, context,
skills, or code, or update model parameters using generated learning
material \citep{reflexion, voyager, r2510_04618, r2505_22954, r2506_10943}.
For example, a skill revision produces a candidate that uses the same
model but contains the revised skill.

\textbf{Candidate feedback.}
$F$ assesses the candidate produced by $P$ and returns the candidate
paired with its feedback $f_t$:
\begin{equation}
F\bigl(\widetilde{\mathcal A}_t;
d_{\mathrm{source}},d_{\mathrm{type}}\bigr)
\longrightarrow\bigl(\widetilde{\mathcal A}_t,f_t\bigr).
\label{eq:rsi-feedback}
\end{equation}
Assessment can involve inspecting the candidate's resources, examining
execution records, or running it to obtain further results. The feedback
$f_t$ provides evidence for the retention decision. \emph{Feedback source},
$d_{\mathrm{source}}$, specifies where the feedback is obtained,
while \emph{Feedback type}, $d_{\mathrm{type}}$, specifies its form,
such as test outcomes, scores, or textual assessments.

\textbf{Retention.}
$R$ uses the current system and the pair returned by $F$ to determine
the configuration retained for the next round:
\begin{equation}
R\bigl(\mathcal A_t,(\widetilde{\mathcal A}_t,f_t);
d_{\mathrm{acceptance}}\bigr)
\longrightarrow\mathcal A_{t+1}.
\label{eq:rsi-retention}
\end{equation}
The \emph{Acceptance criterion}, $d_{\mathrm{acceptance}}$, specifies
which changes to retain, using artifact validation, task results, or
aggregate performance. Accepted changes are incorporated into the
current system; resources that are not replaced remain available.

The settings introduced above give the complete design configuration:
\[
\begin{aligned}
D=\bigl(&d_{\mathrm{mode}},d_{\mathrm{frequency}},d_{\mathrm{scope}},\\
&d_{\mathrm{artifact}},d_{\mathrm{updater}},d_{\mathrm{topology}},\\
&d_{\mathrm{source}},d_{\mathrm{type}},d_{\mathrm{acceptance}}\bigr).
\end{aligned}
\]
Section~\ref{a-unified-taxonomy-of-agent-self-improvement} develops these
nine settings into a taxonomy of existing methods.

In the next round, $\mathcal A_{t+1}$ supports both execution and the
construction of further candidates. A retained skill can guide a later
task and serve as the starting point for another revision. Changes to
the models, instructions, or code used to produce updates can also alter
how subsequent candidates are constructed
\citep{r2505_22954, promptbreeder, r2603_19461}. Successive applications
of $U$ therefore connect retained changes to further self-improvement.

%% file: sections/en/framework.tex
\hypertarget{a-unified-taxonomy-of-agent-self-improvement}{%
\section{A Unified Taxonomy of Agent
Self-Improvement}\label{a-unified-taxonomy-of-agent-self-improvement}}

An RSI method combines choices about what changes, who produces the changes,
when they are made and reused, and how feedback guides their retention.
We propose a unified taxonomy that describes these choices through nine
dimensions organized into four groups. Together, the dimensions define
a method design space. Describing existing methods in this space helps
researchers compare their mechanisms and examine how choices along
different dimensions interact. This analysis can reveal limitations of
current mechanisms and inspire new research ideas for agent improvement.

\subsection{Overview of the unified taxonomy}
\label{dimensions-and-literature-map}

\begin{figure}[tb]
\centering
\includegraphics[width=\linewidth]{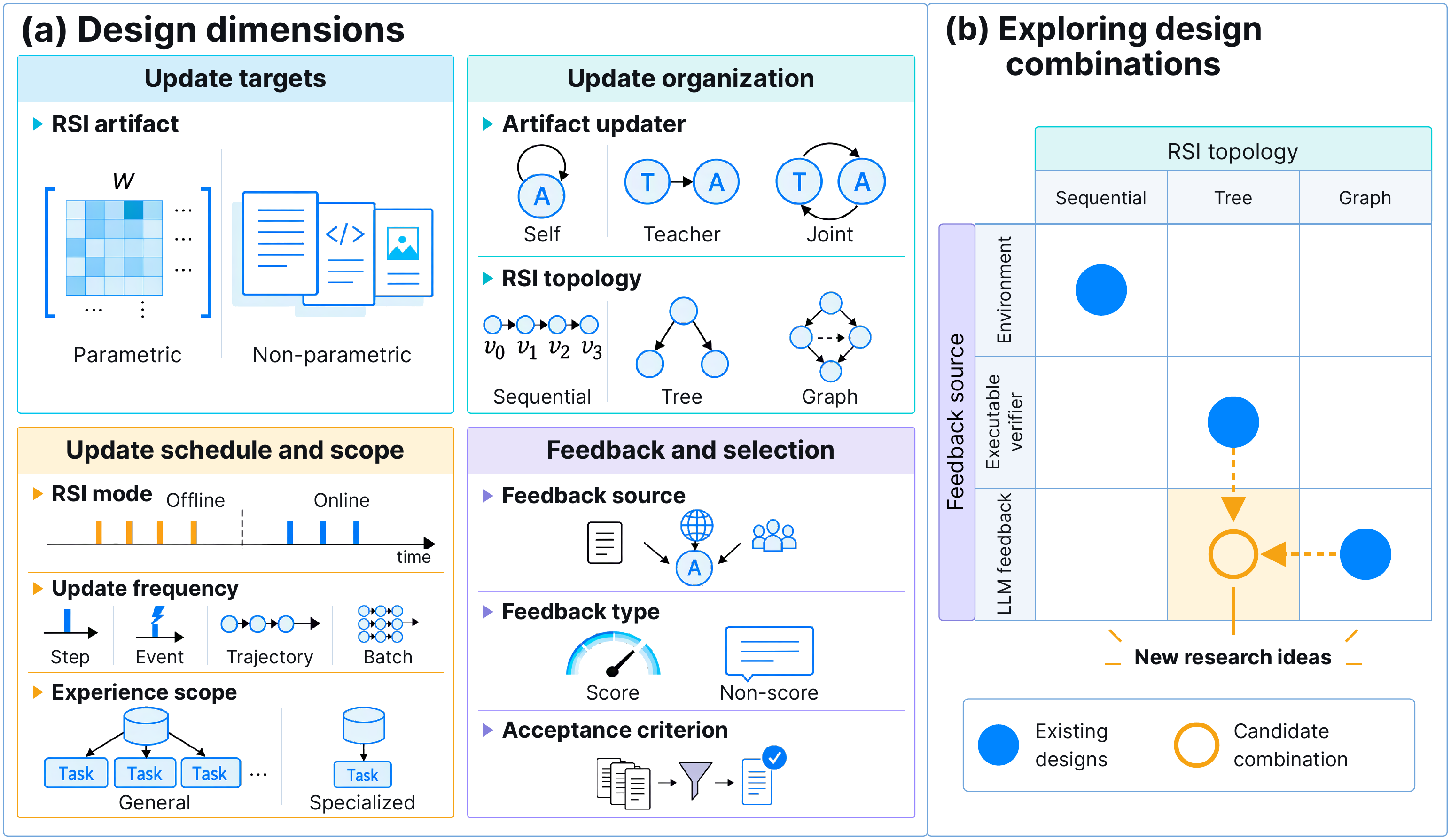}
\caption{The unified taxonomy and its use as a design space. Left: four groups organize nine design dimensions. Right: an illustrative slice locates existing designs and a candidate direction within the same space.}
\label{fig:framework}
\end{figure}

The nine design dimensions are organized into four groups, as shown in
Figure~\ref{fig:framework}.

\begin{enumerate}[label=\arabic*.,font=\bfseries,itemsep=.25\baselineskip]
\item \textbf{Update targets:} RSI artifact.
\item \textbf{Update organization:} Artifact updater and RSI topology.
\item \textbf{Update schedule and scope:} RSI mode, Update frequency, and Experience scope.
\item \textbf{Feedback and selection:} Feedback source, Feedback type, and Acceptance criterion.
\end{enumerate}

\begin{figure}[p]
\centering
\includegraphics[width=\linewidth,height=.93\textheight,keepaspectratio]{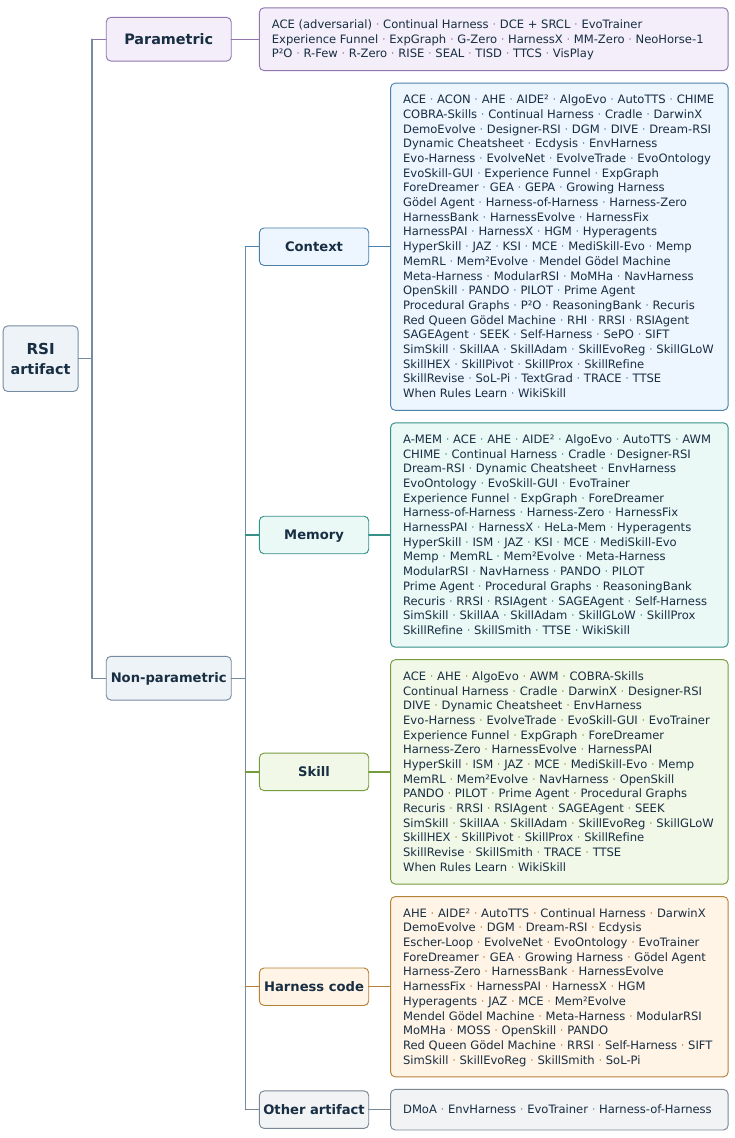}
\caption{Methods grouped by RSI artifact. Papers with multiple labels appear on multiple branches.}
\label{fig:literature-tree}
\end{figure}

Figure~\ref{fig:literature-tree} organizes the reviewed methods by their
update targets. Appendix~\ref{app:catalogue} lists the method
papers, benchmarks, and software systems covered in this survey, together
with their classifications. The following sections define each dimension
and illustrate its choices with representative methods.

\subsection{Update targets}
\label{what-evolves-model-parameters-and-harness-components}

Update targets describe the parts of an agentic system that are modified
and retained for subsequent use. We characterize these targets through
the \emph{RSI artifact} dimension.

\textbf{RSI artifact.} This dimension distinguishes updates to model
weights (\emph{Parametric}) from updates to resources outside those
weights (\emph{Non-parametric}). Non-parametric targets are grouped by
their role in the system. \emph{Context} supplies information to model
calls; \emph{Memory} stores information for later retrieval;
\emph{Skill} packages reusable instructions and supporting resources;
and \emph{Harness code} implements tool operations and coordinates
model calls, tool execution, and control flow. \emph{Other artifact}
covers retained resources outside these categories, such as data
strategies and experiment configurations. A method can update several
artifacts together.

\begin{samepage}
Parametric methods update the language model's weights using training
data and feedback. SEAL generates training material or update
instructions and uses them to fine-tune the model. The updated model's
task performance then serves as a reward for learning to generate more
effective self-edits \citep{r2506_10943}.

\end{samepage}

\begin{samepage}
Context updates change the information supplied to model calls and how
it is organized. They include revising task instructions and selecting,
compressing, or reorganizing observations and interaction history as
long-horizon tasks unfold. GEPA uses execution traces and feedback to
propose revised instructions, then evaluates the candidates before
retaining them \citep{r2507_19457}. ACON optimizes the guidelines used
to compress observations and interaction history. It compares successful
runs with full context against failed runs with compressed context,
then uses these comparisons to revise the guidelines so that compression
preserves information needed for subsequent decisions \citep{r2510_00615}.

\end{samepage}

Memory methods retain information that can be retrieved when relevant.
ReasoningBank extracts strategies from model-judged successful and failed
trajectories. A-MEM organizes interaction records into structured notes
and revises links among related memories \citep{r2509_25140, r2502_12110}.
TTSE separates environmental facts from task-conditioned procedures,
while SAGEAgent distinguishes individual decision records from reusable
rules for acquiring input modalities \citep{r2609_24289, r2607_09521}.
KSI maintains a shared knowledge base whose contents are discussed and
refined by agents across task attempts \citep{r2607_19592}.

Skill methods construct or revise reusable procedures.
TRACE compares successful and failed uses of a skill to revise its
contents, and Evo-Harness compiles execution records into reusable skills
\citep{r2608_22793, r2608_15071}. SkillPivot uses a teacher's
continuation from the same trajectory prefix as a failed attempt to
guide a localized skill edit \citep{r2609_29154}.

Harness code methods change executable behavior. DGM edits the coding
agent's program and retains evaluated versions. Hyperagents allows both
the task agent and the meta-agent that proposes improvements to be
modified \citep{r2505_22954, r2603_19461}. The code target can therefore
include the implementation of task execution or of the update procedure.

Several methods span multiple artifact categories. SkillAdam revises a
skill while retaining a record of earlier editing attempts, and
SkillSmith revises skills together with their executable tools
\citep{r2609_08944, r2606_01314}. AlgoEvo stores evaluated modifications
in persistent cards and consolidates recurring patterns into design
skills, linking Memory with Skill \citep{r2609_15820}. RRSI permits
coordinated changes to harness code, context, memory, and skills around
a frozen model \citep{rrsi}.

\subsection{Update organization}
\label{who-improves-the-agent-and-how-do-versions-interact}

Update organization describes how the system produces successive
versions of itself. It combines two dimensions: \emph{Artifact updater}
identifies who forms a change, and \emph{RSI topology} describes how a
new version builds on earlier versions.

\textbf{Artifact updater.} This dimension distinguishes \emph{Self},
\emph{Teacher}, and \emph{Joint} updates. Self denotes changes produced
by the agent being improved. Teacher denotes changes produced by a
separate updating role using the agent's execution records and feedback.
Joint combines contributions from the agent and a teacher: the agent
proposes or revises a change, and the teacher further refines, curates,
or selects it.

\begin{samepage}
SEAL provides a Self example: the model generates material used to update
its own parameters \citep{r2506_10943}. Recuris assigns skill-memory
revision to a meta-agent, while GEPA uses a reflective optimizer to
revise the evaluated system's instructions; both illustrate Teacher
updates \citep{r2608_24876, r2507_19457}. DGM illustrates Joint updating:
a diagnosis guides a self-modification attempt by the selected coding
agent, and evaluation determines whether the resulting version enters
the archive \citep{r2505_22954}.

\end{samepage}

\textbf{RSI topology.} This dimension describes dependencies between
versions through \emph{Sequential}, \emph{Tree}, and \emph{Graph}
topologies. Sequential improvement repeatedly updates the current
version to form a single chain. A Tree allows archived versions to
branch, with each child derived from one parent. A Graph allows a new
version to draw on retained state, changes, or outcomes from several
historical versions or lineages.

A-MEM illustrates Sequential updates by incorporating new notes and
revising the current memory store \citep{r2502_12110}. DGM selects an
archived agent as a parent and modifies its code to produce a child;
repeatedly branching from archived agents forms a Tree
\citep{r2505_22954}. HGM also maintains a branching archive and uses
descendant performance to guide the choice of promising branches
\citep{hgm}.

Mendel G\"odel Machine illustrates Graph dependencies. It compares the
target agent's execution trace with that of an agent from another lineage
on the same task, then uses the comparison to guide code changes
\citep{r2608_07645}. GEPA's Merge variant draws on complementary changes
from different candidates \citep{r2507_19457}. In both cases, a new
version can inherit useful information from more than one lineage.

\subsection{Update schedule and scope}
\label{when-updates-occur-and-where-they-are-reused}

Update schedule and scope describes when changes are made and which
tasks they are intended to support. Three dimensions specify these
choices: \emph{RSI mode} places improvement relative to system use,
\emph{Update frequency} records when individual updates are triggered,
and \emph{Experience scope} describes the intended range of reuse.

\textbf{RSI mode.} This dimension distinguishes \emph{Offline},
\emph{Online}, and \emph{Offline $\rightarrow$ Online} improvement.
Offline improvement occurs in a separate training or search phase
before evaluation or deployment. Online improvement occurs during
use, so later tasks can use the newly retained state. Offline
$\rightarrow$ Online first builds a resource offline and then continues
to update it during use.

DGM and SkillAdam illustrate Offline improvement by searching for code
or skill revisions before evaluating the resulting system
\citep{r2505_22954, r2609_08944}. A-MEM and ReasoningBank illustrate
Online improvement by incorporating new records or distilled strategies
as tasks are processed \citep{r2502_12110, r2509_25140}. Agent Workflow
Memory studies both offline workflow construction and continued online
induction, including the use of offline workflows to initialize online
improvement \citep{r2409_07429}. ACE separately studies offline playbook
construction and online updating \citep{r2510_04618}.

\textbf{Update frequency.} This dimension describes the trigger or unit
of an individual update through \emph{Step}, \emph{Event},
\emph{Trajectory}, and \emph{Batch}. A Step update follows one
action--observation step. An Event update is triggered by a specified
condition, such as detecting a failure or receiving new information.
A Trajectory update follows a complete task execution, while a Batch
update uses a group of trajectories or candidates.

ReasoningBank extracts memories from completed trajectories, whereas
SkillAdam uses batch-level results to guide skill revisions
\citep{r2509_25140, r2609_08944}. PILOT illustrates how different update
frequencies can coexist. A supervisor can redirect an active worker and
record reusable skills or memory during execution. After a group of runs
finishes, changes from successful runs are merged into the shared state
used by the next iteration \citep{r2608_26530}. These operations correspond
to Event and Batch updates, respectively.

\textbf{Experience scope.} This dimension describes the tasks for which
an updated artifact is intended to be reused. \emph{General} artifacts
support reuse across task types or domains. \emph{Specialized} artifacts
serve a particular instance, topic, or task family. A system can retain
resources with different scopes.

DGM develops a reusable coding harness and studies its transfer between
SWE-bench and Polyglot, illustrating General scope
\citep{r2505_22954}. SkillAdam refines skill documents for particular
task settings, illustrating Specialized scope \citep{r2609_08944}.
AlgoEvo reuses experience both within and across tasks: search records
guide further attempts on the current algorithm-design task, while skills
distilled from multiple tasks guide subsequent algorithm-design tasks
\citep{r2609_15820}.

\subsection{Feedback and selection}
\label{feedback-and-the-retention-of-updates}

Feedback and selection describes how the system obtains guidance for
updates and decides which changes to retain. We separate three
dimensions: \emph{Feedback source} identifies where the guidance comes
from, \emph{Feedback type} describes its form, and
\emph{Acceptance criterion} specifies the basis for retaining a change.

\textbf{Feedback source.} This dimension groups information by its
origin into six categories. A \emph{Train/dev set} provides task
examples and reference outputs; a \emph{Benchmark} supplies results
from a standardized evaluation suite. The \emph{Environment} provides
observations and action outcomes, while an \emph{Executable verifier}
produces test or validation results. \emph{LLM feedback} supplies
model-generated evaluations or suggestions. A \emph{Human} can provide
corrections, preferences, or guidance. An update can use several sources.

DGM uses coding-benchmark results and executable tests to assess revised
agents \citep{r2505_22954}. SIFT adds model-generated pairwise comparisons
between candidate patches to guide parent selection and prioritize
execution-based evaluation \citep{sift}. SkillHEX combines model-generated
failure hypotheses with executable checks: it tests the hypotheses on
cached outputs from skill versions and uses the results to guide later
revisions \citep{r2608_05628}.

\textbf{Feedback type.} This dimension describes the form of feedback,
distinguishing \emph{Score} and \emph{Non-score} information. Score
feedback summarizes an outcome or assessment as a \emph{Binary} result,
such as success or failure, or a \emph{Non-binary} numerical or categorical
evaluation. Non-score feedback supplies content for revising the system:
\emph{Ground truth} provides a reference answer or behavior,
\emph{LLM-as-a-judge} provides a textual assessment, and \emph{Other}
covers observations, execution records, and diagnostics. The type is
determined by the returned information: a numerical model rating is a
Score, while its explanation is Non-score feedback.

GEPA uses execution scores to compare candidates and detailed execution
feedback to guide revisions \citep{r2507_19457}. ReasoningBank judges
whether a trajectory succeeded and extracts lessons from both successful
and failed attempts \citep{r2509_25140}. ACE's reflector analyzes execution
and produces guidance for playbook updates \citep{r2510_04618}. These
methods use both compact assessments and information about the behavior
that produced an outcome.

\textbf{Acceptance criterion.} This dimension describes how a system
decides whether to retain a proposed change. \emph{Artifact validation}
checks whether the modified resource meets requirements such as
successful compilation or interface compatibility. \emph{Instance result}
uses performance on an individual task. \emph{Benchmark score} uses
aggregate performance on a benchmark. \emph{Combined metrics} considers
several measures, such as performance, cost, and stability.
\emph{No validation} denotes direct retention without a separate
acceptance check.

GEPA evaluates candidate instructions before adding successful revisions
to its search population \citep{r2507_19457}. Designer-RSI compares
candidate and reference skills on matched replays with the same upstream
state and retrieved assets. It admits a change when at least one replayed
case improves and no regression is detected on the others
\citep{r2609_22086}. RRSI considers performance gains together with
execution cost and evaluation noise, illustrating Combined metrics
\citep{rrsi}. ACE and A-MEM illustrate No validation by incorporating
curated playbook edits or newly constructed memory records without a
separate performance-based acceptance test
\citep{r2510_04618, r2502_12110}.

\subsection{Ideas from the unified taxonomy}
\label{sec:ideas-taxonomy}

This section illustrates how the unified taxonomy can help generate new
RSI research ideas. Starting from existing methods, we examine the
relationship between relevant dimensions, identify a limitation of their
current designs, and use it to motivate a research direction.

We begin with \emph{Harness code} under the \emph{RSI artifact} dimension.
Code updates can change task execution or the information recorded for
failure diagnosis. HarnessFix can modify model-input logging, context
snapshots, and state-change records \citep{r2606_06324}. EvoTrainer also
revises its diagnostic metrics and analysis procedures when they do not
provide enough information to guide further changes \citep{evotrainer}.

We illustrate this with an agent that can modify its own harness.
During task execution, it uses a web research tool to find papers published
within a given date range, but its requests repeatedly fail. During
self-improvement, the same agent examines its logs, discovers that the
request parameters are missing, and edits its logging code to record them.
When it runs the search task again, the new logs reveal a date format that
the tool does not accept. In the next round of self-improvement, it fixes
the code that prepares the dates. The first edit does not resolve the
search failure, but supplies the evidence needed for the later repair.

Whether the first change is retained, however, depends on the
\emph{Acceptance criterion}. Ecdysis, for example, retains a candidate harness only when its
overall training score exceeds that of the current version
\citep{r2609_11677}. Under such a rule, a logging change that leaves the
score unchanged would be rejected before it could support a later repair.
This reveals a mismatch between the two dimensions: code updates can
improve failure diagnosis, while an acceptance rule based on immediate
task scores may overlook that contribution.

This mismatch motivates a research direction: acceptance criteria that
account for a change's contribution to subsequent self-improvement.
For harness code, the criteria could consider both current task performance
and whether an edit helps the agent diagnose and repair failures in later
updates. The taxonomy guides this idea by
exposing a gap between what the modified resource can contribute and what
the current acceptance \mbox{rule measures.}

%% file: sections/en/connections.tex
\section{Insights from Traditional Deep Learning}
\label{sec:connections}

Traditional deep learning can inform the design of RSI methods. We first examine how existing work uses learning and optimization mechanisms for agent self-improvement. We then explore how further developments of these mechanisms can inspire new research ideas, extending the discussion to broader methods in machine learning and evolutionary computation.

\subsection{Connections in Existing RSI Methods}
\label{sec:connections-existing}

\begin{figure}[!t]
\centering
\includegraphics[width=\linewidth]{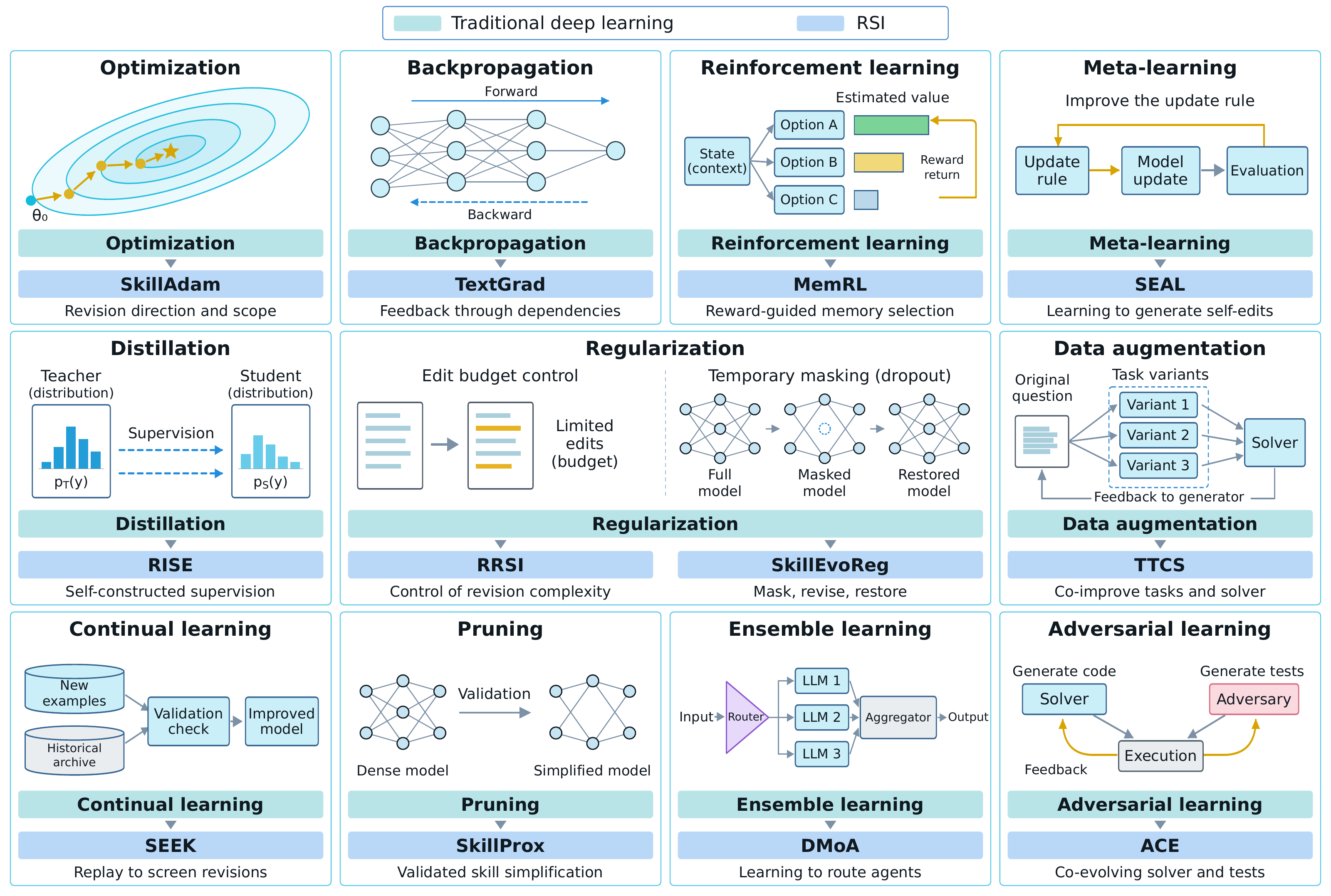}
\caption{Connections between traditional deep learning and RSI, illustrated by representative methods. Each panel links a learning mechanism to its use in RSI; the regularization panel includes both complexity control and temporary masking. Network sketches illustrate the learning principles; SkillEvoReg masks skill instructions, while SkillProx simplifies skill contents.}
\label{fig:transfer}
\end{figure}

Existing RSI methods draw on traditional deep learning to guide how agents learn from feedback and modify themselves. This section examines the problems these learning mechanisms were developed to address and how representative RSI methods apply them to self-improvement. Figure~\ref{fig:transfer} summarizes these connections.

\subsubsection{Optimization}
\label{sec:connections-optimizers}

Optimizers in traditional deep learning determine the direction and magnitude of parameter updates. Adam, for example, uses information accumulated from current and past gradients to compute its parameter updates \citep{adam}. Iterative skill revision in RSI faces a similar problem: deciding what to modify and how much to change at each iteration.

SkillAdam draws on this optimizer design to guide the revision of agent skills \citep{r2609_08944}. It records identified problems, attempted fixes, and their outcomes, allowing later revisions to build on earlier attempts. It also compares task performance before and after a revision to determine how much of the skill may be changed next. When the gains vary substantially across evaluation cases, it reduces the amount of editing allowed in the next iteration; more consistent gains permit broader changes. These mechanisms use the history and results of previous revisions to guide both what to change and how much to change.

\subsubsection{Backpropagation}
\label{sec:connections-backpropagation}

Training a neural network requires using errors in its predictions to adjust parameters across its layers. Backpropagation works backward through the computation graph to calculate how the loss changes with each parameter, providing gradients for parameter updates \citep{backprop}. RSI systems face a related problem: using feedback on a task result to guide changes to individual components.

TextGrad draws on this idea to propagate textual feedback through a system \citep{textgrad}. It represents operations and their input--output relationships as a computation graph, then passes feedback on the final result backward through that graph. At each step, a language model uses the local inputs, outputs, and feedback to generate revision suggestions for upstream variables. An optimizer then uses these suggestions to revise selected variables, such as instructions and code. For RSI, this suggests a way to turn task-level feedback into specific changes by following the steps that produced the result.

\subsubsection{Reinforcement Learning}
\label{sec:connections-rl}

Reinforcement learning uses rewards from interaction to improve decisions. Value-based methods estimate the expected return of available choices and use these estimates to guide selection \citep{suttonbarto}. This principle can also guide memory retrieval: an agent can use task outcomes to learn which memories are useful and adjust its selection on subsequent tasks.

MemRL applies value estimation to this selection process \citep{memrl}. It first retrieves a pool of memories by semantic similarity, then selects from this pool using both relevance and learned utility values. After the agent executes the task, the resulting reward updates the utility values of the memories supplied to the agent. Memories associated with higher rewards are consequently more likely to be selected for later tasks. With the language model's parameters fixed, feedback improves how the agent selects from its accumulated records.

\subsubsection{Meta-Learning}
\label{sec:connections-meta}

Meta-learning uses performance after learning to improve how learning is carried out. For example, it can optimize a model's initialization or learn a parameter-update rule by evaluating how well the model performs after training \citep{maml,learnopt}. In RSI, this principle can guide the production of training data and update instructions: their quality can be assessed through the learning they produce.

SEAL implements this idea by training a model to generate its own self-edits \citep{r2506_10943}. Given new information or a task, the model produces training materials or instructions for updating its parameters. The proposed update is applied through fine-tuning, and the updated model is evaluated on the relevant task. Its performance then serves as a reward for reinforcement learning of the self-edit generator. This creates two connected learning processes: one updates the model using generated materials, and the other improves how those materials and update instructions are produced.

\subsubsection{Distillation}
\label{sec:connections-distillation}

Knowledge distillation transfers behavior from a teacher model to a student by training the student to match the teacher's output probabilities \citep{distillation}. These probabilities provide guidance about alternative outputs beyond a single target answer. A self-improving system can use this mechanism while also constructing and updating the source of its supervision.

RISE combines distillation with reinforcement learning using verifiable rewards \citep{rise}. It compares the model before and after a reinforcement learning update, then constructs a teacher by extending the observed change farther in the same direction. This extrapolation can operate on model weights or output log-probabilities. The student is trained to match the resulting teacher's token probabilities. As reinforcement learning and distillation alternate, each new training update provides the basis for another teacher. RISE thus incorporates teacher construction into the improvement process, using the model's own training trajectory to produce supervision for subsequent learning.

\subsubsection{Regularization}
\label{sec:connections-regularization}

Regularization aims to reduce overfitting and improve generalization beyond the training data. Common approaches constrain model complexity or introduce perturbations during learning. $L_0$ regularization penalizes the number of nonzero parameters to encourage sparsity \citep{l0regularization}. $L_1$ regularization encourages sparsity by penalizing the sum of absolute parameter values, while $L_2$ regularization discourages large weights by penalizing the sum of squared parameter values \citep{deeplearning}. Dropout introduces perturbations by temporarily masking units during training, reducing reliance on particular combinations of features \citep{dropout}. RSI faces a similar difficulty when repeated harness or skill revisions fit the tasks used for improvement but transfer poorly to new tasks.

RRSI applies complexity control to harness evolution \citep{rrsi}. It gradually reduces the number of independent edits allowed in each proposal, limiting how many changes can be introduced in response to a round of feedback. Components that repeatedly fail to yield gains are marked for removal. Candidate selection also requires increased execution cost to be justified by improved task performance.

SkillEvoReg adapts dropout to skill revision \citep{r2609_30861}. It randomly masks instructions or procedural blocks in the current skill. The agent executes a training task with this partial skill, and the updater uses the resulting execution to propose a revision. The system then restores the masked content and integrates the proposed changes into the complete skill. The aim is to reduce the system's reliance on particular existing instructions when revising a skill.

\subsubsection{Data Augmentation}
\label{sec:connections-augmentation}

Data augmentation expands training data by constructing variants of existing examples. In image classification, for example, crops and horizontal reflections provide additional views of labeled images \citep{alexnet}. For a self-improving agent, task variants can similarly provide further practice on the reasoning or operations required by the problems it encounters.

TTCS uses a question synthesizer to generate variants of test questions \citep{r2601_22628}. It changes problem objects, settings, or constraints while aiming to preserve the underlying reasoning structure. The solver trains on both original and generated questions, using agreement among sampled answers as a learning signal. Solver feedback also updates the synthesizer. The system thus learns from the additional data while improving how those data are generated.

\subsubsection{Continual Learning}
\label{sec:connections-continual}

Continual learning concerns learning from successive tasks or data while retaining earlier capabilities. Rehearsal reuses past examples during later learning; iCaRL combines this with distillation to preserve previously learned classes \citep{icarl}. Agent self-improvement raises the same retention problem when a new skill revision disrupts tasks that the agent could previously complete.

SEEK uses historical replay to screen such revisions in an industrial search evaluator \citep{r2609_29803}. Reviewed errors guide changes to the operational instructions in a skill bank. Each candidate is then checked on emerging cases and a fixed set of historical examples. Acceptance requires improvement on the new cases while limiting regression on the historical set. The retained examples serve as a check on successive skill updates, extending the role of past data from parameter training to the acceptance of reusable instructions.

\subsubsection{Pruning}
\label{sec:connections-pruning}

Pruning removes network weights or structures that contribute little to the desired behavior, often followed by further training \citep{deepcompression}. For agents, repeated revisions can similarly accumulate instructions and resources whose contribution no longer justifies keeping them.

SkillProx implements a related simplification step after skill revision \citep{skillprox}. It measures the utility of skill sections and reference groups by removing them and rerunning validation cases. Low-utility units become candidates for deletion, demotion, or consolidation. Each proposed simplification must reduce complexity, preserve structural validity, and pass performance checks. The resulting skill becomes more compact through measured changes to its contents. The method also draws on proximal optimization by separating performance-oriented revision from subsequent complexity reduction.

\subsubsection{Ensemble Learning}
\label{sec:connections-ensemble}

Ensemble learning combines predictions from multiple learners to use their complementary strengths \citep{ensemblelearning}. In agentic systems, this connection concerns how several agents contribute to a solution and how their participation is learned.

DMoA learns a router that selects agents at successive reasoning steps, combines their responses into the next step's context, and uses a summarizer to produce the final answer \citep{dmoa}. Its test-time training setting first runs all candidate agents on an initial set of queries and collects predictive entropy as supervision. It then updates the router and uses the learned selection on subsequent queries. This protocol illustrates how a system can learn to coordinate an ensemble from its own outputs while keeping the underlying language models fixed.

\subsubsection{Adversarial Learning}
\label{sec:connections-adversarial}

Adversarial learning trains interacting models with competing objectives. In generative adversarial networks (GANs), a discriminator learns to distinguish real data from generated samples, while a generator learns to produce samples that the discriminator classifies as real \citep{gan}. As the two models are updated, each learns against a changing opponent. In RSI, a related interaction can couple task solving with the search for failures in current solutions.

The code-generation framework ACE illustrates this interaction through a solver and an adversary \citep{ace_adversarial}. The solver generates candidate programs, while the adversary generates test inputs intended to trigger execution failures. The system evaluates the programs using both existing correctness tests and the generated inputs, then fine-tunes the solver on selected programs. Execution results also provide preferences for training the adversary: tests are favored when they cause some candidates to fail while others still execute successfully. Each new round uses the updated solver and adversary, allowing failure discovery and code generation to improve through continued interaction.

\subsection{From Established Methods to New RSI Ideas}
\label{sec:ideas-dl}

\begin{figure}[!htbp]
\centering
\includegraphics[width=\linewidth]{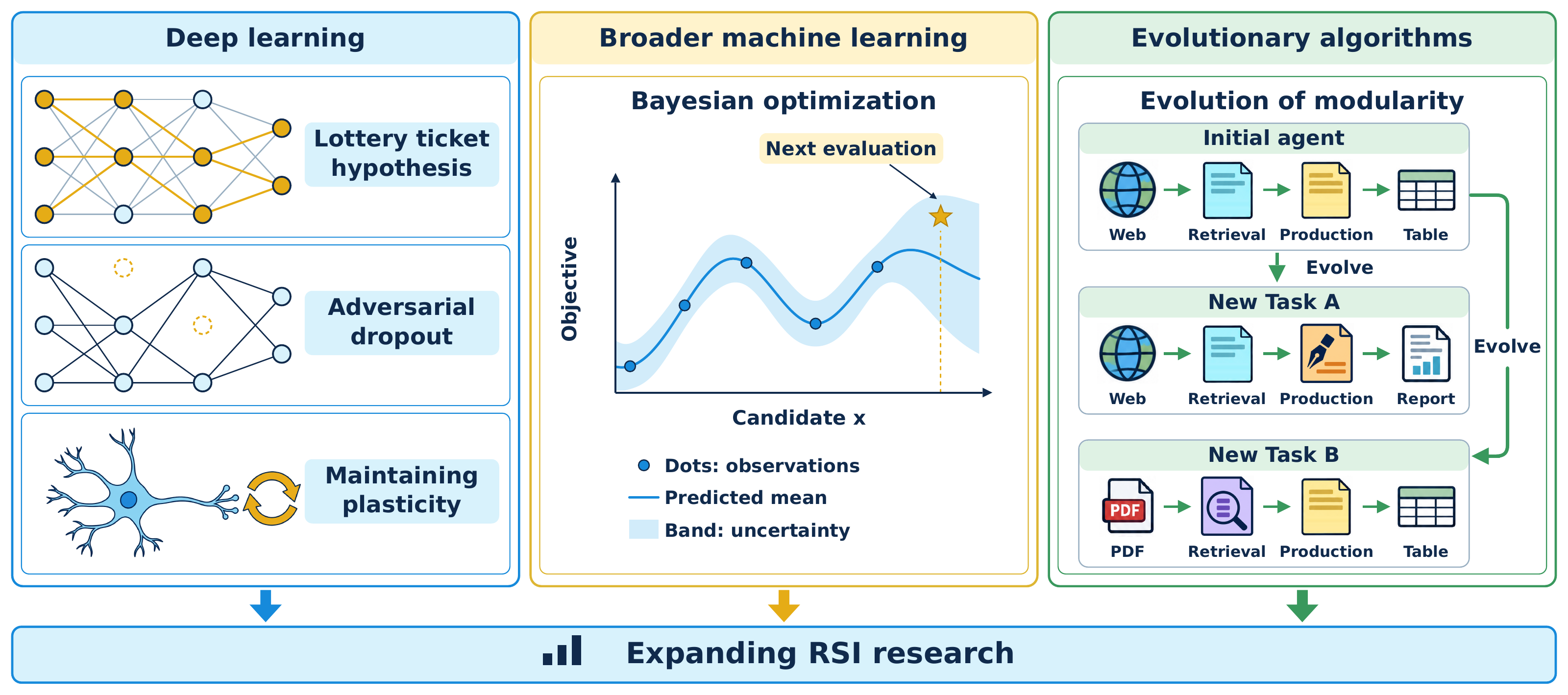}
\caption{Broadening the sources of ideas for RSI. Left: the lottery ticket hypothesis, adversarial dropout, and maintaining plasticity in deep learning. Center: Bayesian optimization. Right: evolution of modularity, illustrated by revising production or retrieval modules to adapt to new tasks. Bottom arrows indicate how these sources can inspire RSI research directions.}
\label{fig:learning-directions}
\end{figure}

This section explores how further developments of learning methods can inspire new RSI research ideas. We begin with deeper developments in traditional deep learning, then broaden the discussion to other machine learning methods and evolutionary algorithms. Figure~\ref{fig:learning-directions} places these sources within a wider landscape of connections to RSI.

\subsubsection{Further Developments in Traditional Deep Learning}
\label{sec:ideas-deeper-dl}

We first deepen the connections established in Section~\ref{sec:connections-existing}, considering more specific mechanisms from traditional deep learning.

\textbf{The Lottery Ticket Hypothesis.} SkillProx uses pruning to simplify skills while preserving their task performance, as discussed in Section~\ref{sec:connections-pruning}. The lottery ticket hypothesis takes pruning a step further: a large network can contain a small subnetwork that, when trained from its original initial weights, reaches comparable accuracy in a similar number of iterations \citep{lotteryticket}.

For RSI, the lottery ticket hypothesis suggests examining which resources an agent needs at the start of self-improvement. A system with an extensive collection of skills and code might need only a subset of its early resources to develop comparable capabilities through subsequent revisions. If such a starting point exists, it could enable agents to improve effectively with fewer resources.

\textbf{Adversarial Dropout.} Dropout randomly masks units during training. Adversarial dropout develops this idea by seeking bounded changes to the mask that most increase prediction error, then training the network to cope with them \citep{adversarialdropout}. SkillEvoReg already applies random masking to agent skills during execution and revision \citep{r2609_30861}. This connection suggests using targeted masking to expose strong dependencies in skills and encourage alternative procedures.

For example, a masking agent could identify the skill content on which a web research agent strongly depends. If hiding a template tailored to one website causes extraction to fail, it could use that disruption to encourage the system to explore other extraction procedures, such as locating information through page headings and nearby text. As the skills improve, the masking agent could seek further masks that expose weaknesses, providing new challenges for subsequent self-improvement.

\looseness=-1
\textbf{Maintaining Plasticity.} Section~\ref{sec:connections-continual} describes how past examples help retain earlier capabilities. Continual learning also requires plasticity: the ability to learn new behavior through further training. This ability can decline as a network learns successive tasks. Continual backpropagation addresses this problem by selectively reinitializing a small fraction of low-utility hidden units during training, allowing them to learn new features \citep{continualbackprop}.

An agent that improves itself through parameter updates can use its own verified task solutions as training data. However, prolonged training may gradually reduce the model's ability to learn new ways of solving tasks. Drawing on continual backpropagation, the system could reinitialize a small fraction of low-utility neurons during training, then resume training so these neurons can learn new features. This could help the agent maintain its ability to learn from new tasks across successive rounds of self-improvement.

\subsubsection{Insights from Broader Machine Learning}
\label{sec:ideas-broader-ml}

Broader machine learning methods can also inspire new research ideas for RSI. We illustrate this with a concrete example below.

\textbf{Bayesian Optimization.} Bayesian optimization uses a probabilistic model of previous evaluations to guide the choice of subsequent evaluations. Freeze-Thaw Bayesian Optimization extends this approach to partially completed training runs \citep{freezethaw}. It uses a Gaussian process to predict future performance from observed learning curves and estimate the uncertainty of these predictions. This information guides decisions to continue a training run, pause it, or resume a previously paused run.

An RSI system exploring several branches of self-improvement could use this mechanism to decide which branches to develop further. Following Freeze-Thaw, a lightweight Gaussian process predictor could use the performance recorded after successive self-modifications to estimate what additional rounds might achieve. For example, one branch of a web research agent might score well but make little recent progress, while another starts lower and continues to improve. Predicted performance and its uncertainty could guide which branch to continue, pause, or resume. Paused branches would retain the agent state and revision history needed to continue their self-improvement. As further revisions are evaluated, their results would update the predictor and guide the allocation of subsequent improvement rounds.

\subsubsection{Insights from Evolutionary Algorithms}
\label{sec:ideas-evolutionary}

Evolutionary algorithms can likewise inspire new research ideas for RSI. We illustrate this with a concrete example below.

\textbf{Evolution of Modularity.} Changing one part of a skill can disturb another part that already works. Separating procedures that can change independently could make later revisions easier. ModularRSI, for example, improves predefined harness modules separately \citep{r2609_14857}. Evolutionary experiments suggest that useful divisions can also emerge as task requirements change. \citet{evolvingmodularity} repeatedly switched between goals made from different combinations of shared subgoals. The networks evolved modules that could be reused when the goal changed.

For RSI, consider a research agent that initially uses one skill to retrieve webpages and produce a comparison table. Later tasks sometimes require a written report from the same webpages, and sometimes require the same table from PDF documents. During self-improvement, the updater could identify which procedures need changing in each case: a new source requires different retrieval steps, while a new output format requires different reporting steps. It could then separate these procedures into reusable skills. The agent could subsequently learn to read a new document type while retaining its report-writing skill, or improve report writing while retaining its retrieval skill.

%% file: sections/en/evaluation.tex
\section{Evaluation and Benchmarks}
\label{sec:evaluation}

RSI evaluation examines how self-improvement changes an agent's ability to
solve tasks. A benchmark defines a \emph{problem setup} through the tasks it
provides, the feedback available to the system, the updates it allows, and
the way it assesses the resulting changes. These choices determine which
aspects of self-improvement the evaluation measures.

A central distinction is how improvement is combined with testing. We
organize benchmarks and evaluation settings into \emph{Online},
\emph{Offline}, and \mbox{\emph{Offline $\rightarrow$ Online}} according to
whether the system improves during evaluation, before evaluation, or in
both phases. Within each setting, we compare task organization, available
feedback, and evaluation goals. Table~\ref{tab:evaluation-overview}
compares representative studies, and Appendix~\ref{app:catalogue}
provides the full benchmark catalogue.

\begin{table}[!htb]
\centering\small
\setlength{\tabcolsep}{4pt}
\renewcommand{\arraystretch}{1.10}
\caption{Comparison of representative RSI benchmarks and evaluation settings.}
\label{tab:evaluation-overview}
\begin{tabular}{@{}>{\raggedright\arraybackslash}p{0.16\dimexpr\linewidth-8\tabcolsep\relax}>{\raggedright\arraybackslash}p{0.19\dimexpr\linewidth-8\tabcolsep\relax}>{\raggedright\arraybackslash}p{0.25\dimexpr\linewidth-8\tabcolsep\relax}>{\raggedright\arraybackslash}p{0.20\dimexpr\linewidth-8\tabcolsep\relax}>{\raggedright\arraybackslash}p{0.20\dimexpr\linewidth-8\tabcolsep\relax}@{}}
\toprule
Study & Domain & Evaluation setup & Feedback for improvement & Evaluation focus \\
\midrule
\multicolumn{5}{@{}l}{\textbf{Online}} \\
Evo-Memory \citep{r2511_20857} & Reasoning, coding, and interactive tasks. & Streams with varying task difficulty and order. & Correctness signals after execution. & Performance gains from using earlier task records. \\[2pt]
SkillFlow \citep{r2604_17308} & Finance, healthcare, documents. & Related tasks in increasing difficulty. & Execution traces and verifier feedback. & Skill acquisition, repair, and reuse. \\[2pt]
AgentStream \citep{r2608_00155} & Web research, coding, and tool use. & Isolated, sequential, or interleaved domains. & Execution results and self-evaluation. & Transfer and interference across domains. \\[2pt]
PATH-Bench \citep{r2608_01149} & Code generation and tool use. & Probes repeated after helpful or interfering examples. & Execution records between probes. & Transfer and forgetting. \\
\midrule
\multicolumn{5}{@{}l}{\textbf{Offline}} \\
MemoryBench \citep{r2510_17281} & QA and writing: general, legal, academic. & Separate training and test examples. & Simulated user comments and actions. & Generalization to unseen examples. \\[2pt]
GDPevo \citep{r2608_03764} & Enterprise workflows. & Rules learned separately are combined in held-out workflows. & Reference answers, grader scores, or observations. & Transfer to new combinations of learned rules. \\[2pt]
HarnessOpt-Bench \citep{r2608_06301} & Document QA, search, and terminal tasks. & Select a candidate harness, then test it on held-out examples. & Development results and traces, aggregate validation scores. & Held-out gain over the seed harness. \\[2pt]
S3Gym \citep{r2608_31100} & Text-based games. & Exploration followed by held-out evaluation. & Environment observations and self-judgments. & Self-judgment accuracy and held-out gains. \\
\midrule
\multicolumn{5}{@{}l}{\textbf{Offline $\rightarrow$ Online}} \\
AWM \citep{r2409_07429} & Web navigation on Mind2Web. & Training-derived workflows followed by updates during testing. & Demonstrations and model success judgments. & Transfer to new tasks, websites, and domains. \\[2pt]
Mem$^2$Evolve \citep{r2604_10923} & QA, math, planning, and interactive tasks. & Memory from prior examples, updated during testing. & Model success judgments and critiques. & Cross-benchmark transfer and continued learning. \\
\bottomrule
\end{tabular}
\end{table}

\subsection{Online evaluation}
\label{sec:evaluation-online}

Online evaluation allows a system to update itself during task execution
and use retained changes in subsequent work. It examines whether the
system improves through continued use and how its performance changes as
new tasks arrive. The relationships among tasks and the feedback they
provide shape the opportunities for improvement.

To examine reuse across tasks, Evo-Memory organizes reasoning,
question-answering, and interactive tasks into streams
\citep{r2511_20857}. The system retrieves memory before solving a task and
updates it after the interaction, using records of its execution and
correctness feedback. Evaluating later tasks with the accumulated memory
measures how well the system uses information from earlier interactions.
The benchmark also varies task difficulty order and feedback to examine
how these conditions affect learning.

Tasks that share a procedure provide a more specific test of reuse.
SkillFlow constructs task families around a common workflow
\citep{r2604_17308}. Agents start without skills, solve the tasks
sequentially, and use execution traces and feedback based on scoring
rubrics to create or repair skill documents. The updated library remains
available for later tasks in the family. ContinualSkillBench arranges
related tasks by difficulty and opportunities for skill reuse
\citep{r2608_03874}. It compares independent task execution, accumulated
context, and explicit skill maintenance. These settings examine whether
what is learned from one task helps solve related tasks, and how the form
of retained information affects that transfer.

Changing the task sequence makes transfer and retention observable.
AgentStream organizes agent tasks into isolated, sequential, and
interleaved domain settings \citep{r2608_00155}. This allows comparisons
between learning within one domain and carrying updates across domains.
PATH-Bench places recurring probe tasks among histories designed to help
or interfere with them \citep{r2608_01149}. Revisiting the probes reveals
whether later updates preserve earlier capabilities or cause forgetting.
The sequence therefore forms part of the problem being evaluated:
performance on a task depends on what the agent encountered before it.

Online evaluation can also track successive attempts at an ongoing
research objective. In RSIBench-Data, a research agent repeatedly revises
training-data strategies for a target model \citep{r2607_25886}. A fixed
training and evaluation pipeline returns results after each experiment,
which inform the next revision. This setting examines how an agent uses
experimental feedback to improve its research decisions over a sequence
of attempts.

\subsection{Offline evaluation}
\label{sec:evaluation-offline}

Offline evaluation separates improvement from testing. The system first
learns or searches on training tasks, then solves held-out tasks with
further self-improvement updates disabled. The evaluation asks whether
the changes acquired during improvement generalize. The relationship
between training and test tasks specifies the kind of transfer being
examined.

MemoryBench tests transfer from feedback on training tasks to separate
test tasks \citep{r2510_17281}. A simulated user provides feedback on the
system's training responses, and a separate evaluation process scores
its test responses. The tasks span several domains, languages, and
input--output formats. This setup examines whether feedback can support
learning that is useful beyond the examples that produced it.

Some benchmarks explicitly construct the connection between learning and
test tasks. GDPevo distributes business rules across training tasks and
recombines them in held-out enterprise workflows \citep{r2608_03764}.
The system must apply rules it has learned in combinations it has not
practiced. EvoAgentBench links training and test tasks through shared
procedural requirements \citep{r2607_05202}. These designs make the
expected transfer concrete by specifying which procedures learned during
improvement should help with a later test task.

Offline evaluation also tests whether harness revisions improve
performance on unseen tasks. HarnessOpt-Bench supplies an initial harness
and feedback for revising it, then compares the selected version with
the initial one on separate test tasks \citep{r2608_06301}. HarnessDev
evaluates both harness construction and subsequent improvement, testing
the resulting programs on unseen tasks after their code is frozen
\citep{r2609_01437}. To reveal the effects of harness revision,
Evo-Bench prioritizes tasks whose scores reflect differences in harness
quality, then divides them into separate sets for improvement and final
testing \citep{r2608_09096}.

Offline evaluation also examines learning from exploration. In S3Gym,
an agent explores text-based games and
judges its own behavior before stricter held-out assessment. It can carry
information forward through interaction history, summary memory, or
parameter updates \citep{r2608_31100}. This setup tests how information acquired through
exploration supports later task solving.

\subsection{Offline-to-online evaluation}
\label{sec:evaluation-hybrid}

Offline-to-online evaluation begins with resources acquired during an
offline phase and allows further updates during evaluation. It examines
how initial preparation supports subsequent tasks and how the system
continues learning from new interactions. Both the source of the initial
resources and the relationship between earlier and later tasks are
therefore part of the problem setup.

Agent Workflow Memory (AWM) provides an example through its combined
evaluation on Mind2Web \citep{r2409_07429}. It first induces reusable web
workflows from training examples, then adds workflows learned while
processing test tasks. The cross-task, cross-website, and cross-domain
splits vary how far those tasks depart from the training distribution.
Comparing the combined setting with offline-only and online-only settings
examines how previously learned workflows interact with those acquired
during continued use.

The memory-initialization experiments in Mem$^2$Evolve study a related
setup \citep{r2604_10923}. The system first builds memory from a subset
of benchmark tasks, then continues to use and update memory while
processing the remaining tasks. Another setting initializes memory
from GAIA before moving to other benchmarks. These arrangements examine
how resources learned within the same task family or elsewhere support
continued improvement.

\subsection{Toward unified evaluation}
\label{sec:evaluation-unified}

Evaluation settings differ across studies. Online, offline, and
offline-to-online evaluation allow self-improvement at different stages
and assess different aspects of improvement. Even within the same mode,
studies can differ in the benchmarks they use, the order of evaluation
examples, and the feedback they provide. For example, what a system learns
from earlier examples can help it solve later ones, while reference
answers, test results, and model-generated assessments provide different
information for revision. Comparing the gains reported by different
methods therefore requires considering these evaluation conditions.

Establishing shared benchmarks and evaluation protocols is important for
RSI research. Within each evaluation mode, common task organization,
feedback conditions, and scoring rules would make comparisons more
informative. Reporting task performance together with the gain over the
same system before improvement would further clarify the contribution of
self-improvement. This common basis would help identify designs that
support generalization, continued improvement, and retention of acquired
capabilities.

%% file: sections/en/conclusion.tex
\section{Conclusion}
\label{sec:conclusion}

This survey systematically reviews existing RSI methods through a unified
taxonomy and examines how some of these methods draw on traditional deep
learning. The taxonomy places existing methods in a shared high-dimensional
design space, providing a basis for understanding and comparing them.
Examining individual dimensions and their interactions can reveal limitations
of current mechanisms and suggest new design directions. Through concrete
examples, we explore how more specific theories and mechanisms within
established learning methods can inspire new RSI research ideas, and extend
the search for inspiration beyond traditional deep learning.

%% file: sections/en/catalogue.tex
\begin{landscape}
\section{Literature Catalogue}
\label{app:catalogue}

This appendix collects the recursive self-improvement methods, benchmarks,
and software systems discussed in this survey. The method catalogue applies the unified taxonomy
from Section~\ref{a-unified-taxonomy-of-agent-self-improvement}; the benchmark
catalogue summarizes the evaluation settings reviewed in
Section~\ref{sec:evaluation}.

\subsection{Method papers}

Table~\ref{tab:coordinates} records each method's choices across the nine
dimensions and briefly describes its update mechanism. Multiple labels
capture the resources and variants studied in each paper.

\input{tables/methods.tex}

\Needspace{16\baselineskip}
\subsection{Benchmarks}

Table~\ref{tab:benchmarks} summarizes the evaluation settings of the reviewed benchmarks.
It records the resources being updated, how improvement and testing are
organized, the feedback available for improvement, and the reported measures.
Task performance includes success rate, reward, and output quality; gain
measures improvement over a reference system. Cost and time cover resource
use and elapsed time, respectively.

\input{tables/benchmarks.tex}

\Needspace{12\baselineskip}
\subsection{Software systems}

\begingroup\footnotesize\setlength{\tabcolsep}{4pt}\renewcommand{\arraystretch}{1.15}
\hypersetup{urlcolor=black}
\begin{longtable}{@{}>{\raggedright\arraybackslash}p{0.14\dimexpr\linewidth-4\tabcolsep\relax}>{\raggedright\arraybackslash}p{0.45\dimexpr\linewidth-4\tabcolsep\relax}>{\raggedright\arraybackslash}p{0.41\dimexpr\linewidth-4\tabcolsep\relax}@{}}
\caption{Software systems for agent self-improvement.}\label{tab:software}\\
\toprule
System & Main functionality & Repository\\\midrule
Proteus \citep{proteus} & Evolves agent harnesses, records version snapshots, and measures changes. & \url{https://github.com/proteus-evolve/Proteus}\\[4pt]
Reef \citep{reef} & Connects execution records and feedback to updates, evaluation, and version deployment. & \url{https://github.com/Human-Agent-Society/reef}\\[4pt]
ReMe \citep{reme} & Stores, retrieves, and maintains agent memories for reuse across tasks. & \url{https://github.com/agentscope-ai/ReMe}\\
\bottomrule
\end{longtable}\endgroup
\end{landscape}

%% file: tables/methods.tex
\begingroup\footnotesize\setlength{\tabcolsep}{3pt}\renewcommand{\arraystretch}{1.08}
\setlength{\emergencystretch}{1em}
\begin{longtable}{@{}>{\raggedright\arraybackslash}p{0.14\dimexpr\linewidth-12\tabcolsep\relax}>{\raggedright\arraybackslash}p{0.06\dimexpr\linewidth-12\tabcolsep\relax}>{\raggedright\arraybackslash}p{0.12\dimexpr\linewidth-12\tabcolsep\relax}>{\raggedright\arraybackslash}p{0.12\dimexpr\linewidth-12\tabcolsep\relax}>{\raggedright\arraybackslash}p{0.18\dimexpr\linewidth-12\tabcolsep\relax}>{\raggedright\arraybackslash}p{0.12\dimexpr\linewidth-12\tabcolsep\relax}>{\raggedright\arraybackslash}p{0.26\dimexpr\linewidth-12\tabcolsep\relax}@{}}
\caption{Method papers and their classifications under the unified taxonomy.}\label{tab:coordinates}\\
\multicolumn{7}{@{}p{\linewidth}@{}}{P: parameters; H: harness code; C: context; M: memory; S: skills; O: other artifacts. Off/On: offline/online.}\\[3pt]
\toprule
\textbf{Method} & \textbf{Targets} & \textbf{Organization}\newline Updater / topology & \textbf{Schedule and scope}\newline Mode / frequency / scope & \textbf{Feedback}\newline Source / type & \textbf{Acceptance} & \textbf{Update mechanism} \\
\midrule\endfirsthead
\multicolumn{7}{@{}l}{\tablename\ \thetable\ (continued)}\\[3pt]
\multicolumn{7}{@{}p{\linewidth}@{}}{P: parameters; H: harness code; C: context; M: memory; S: skills; O: other artifacts. Off/On: offline/online.}\\[3pt]
\toprule
\textbf{Method} & \textbf{Targets} & \textbf{Organization}\newline Updater / topology & \textbf{Schedule and scope}\newline Mode / frequency / scope & \textbf{Feedback}\newline Source / type & \textbf{Acceptance} & \textbf{Update mechanism} \\
\midrule\endhead
\midrule\multicolumn{7}{r@{}}{\textit{Continued on the next page.}}\\\endfoot
\bottomrule\endlastfoot
\textbf{A-MEM}\newline\citep{r2502_12110} & M & Teacher\newline Sequential & On\newline Event\newline General & LLM feedback\newline Other & No validation & Turns each new memory into a structured note and lets it revise summaries, attributes, and links of related historical memories in a Zettelkasten-style graph. \\[3pt]
\textbf{ACE}\newline\citep{r2510_04618} & C, M, S & Teacher\newline Sequential & Off, On\newline Trajectory, Batch\newline General, Specialized & Train/dev set, Benchmark, Environment, Executable verifier, LLM feedback\newline Binary, Non-binary, Ground truth, LLM-as-a-judge, Other & No validation & Turns execution feedback into incremental updates to a structured playbook, preserving useful strategies while revising and deduplicating experience for later tasks. \\[3pt]
\textbf{ACE (adversarial)}\newline\citep{ace_adversarial} & P & Joint\newline Sequential & Off\newline Batch\newline Specialized & Train/dev set, Executable verifier\newline Binary, Non-binary & No validation & Updates solver and adversary adapters across rounds, using selected programs to train the solver and execution-derived preferences to train adversarial test generation. \\[3pt]
\textbf{ACON}\newline\citep{r2510_00615} & C & Teacher\newline Sequential & Off\newline Batch\newline Specialized & Train/dev set, Environment, Executable verifier, LLM feedback\newline Binary, Non-binary, LLM-as-a-judge, Other & Benchmark score, Combined metrics & Contrasts successful full-context runs with failed compressed-context runs to revise compression guidelines, selecting candidates by task success and context cost. \\[3pt]
\textbf{AHE}\newline\citep{r2604_25850} & H, C, M, S & Teacher\newline Sequential & Off\newline Batch\newline General & Benchmark, Executable verifier, LLM feedback\newline Binary, Non-binary, LLM-as-a-judge, Other & Artifact validation, Benchmark score & Distills batches of coding-agent trajectories into layered diagnostics, edits file-level harness components, and uses next-round task outcomes to verify or roll back each predicted change. \\[3pt]
\textbf{AIDE$^2$}\newline\citep{r2609_26457} & H, C, M & Teacher\newline Graph & Off\newline Batch\newline General & Train/dev set, Environment, Executable verifier\newline Non-binary, Other & Benchmark score & Rewrites an AI research agent's code using prior candidate evaluations, retaining versions that improve research performance under fixed task budgets. \\[3pt]
\textbf{AlgoEvo}\newline\citep{r2609_15820} & C, M, S & Self\newline Graph & On\newline Event, Batch\newline General, Specialized & Train/dev set, Environment, Executable verifier, LLM feedback\newline Non-binary, Other & Instance result & Turns algorithm-search feedback into persistent experience cards and consolidates patterns supported by multiple tasks into reusable design skills for subsequent discovery. \\[3pt]
\textbf{AutoTTS}\newline\citep{r2605_08083} & H, C, M & Teacher\newline Graph & Off\newline Batch\newline General & Train/dev set, Executable verifier\newline Non-binary, Other & Combined metrics, Benchmark score & An LLM agent discovers test-time scaling controllers for other LLMs inside an environment built from pre-collected reasoning trajectories and probe signals, using execution-trace feedback to revise controller programs that then transfer to held-out benchmarks and model scales. \\[3pt]
\textbf{AWM}\newline\citep{r2409_07429} & M, S & Teacher\newline Sequential & On, Off, Off$\to$On\newline Trajectory, Batch\newline Specialized & Train/dev set, Environment\newline Ground truth, Other & No validation & Induces reusable workflows from demonstrations or the agent's own completed web trajectories and retrieves them to guide later tasks. \\[3pt]
\textbf{CHIME}\newline\citep{r2609_02074} & C, M & Self\newline Sequential & On, Off\newline Trajectory\newline Specialized & Train/dev set, Environment, Executable verifier, LLM feedback\newline Binary, Non-binary, Other & Instance result & Separates planning and execution memories, attributes task outcomes before storing experience, and updates the relevant bank to guide later tasks without changing model weights. \\[3pt]
\textbf{COBRA-Skills}\newline\citep{r2609_11682} & C, S & Teacher\newline Graph & Off\newline Batch\newline Specialized & Train/dev set, Environment, LLM feedback\newline Binary, Non-binary, Other & Benchmark score & Uses a contextual bandit to allocate skill evaluations, then evolves a population through pruning, mutation, and crossover using observed task rewards. \\[3pt]
\textbf{Continual Harness}\newline\citep{r2605_09998} & P, H, C, M, S & Self, Teacher\newline Sequential & On\newline Step, Batch\newline Specialized & Environment, LLM feedback\newline Non-binary, LLM-as-a-judge, Other & No validation & Refines prompts, subagents, skills, and memory every few environment steps without resetting the episode, and optionally co-learns model weights from process-reward-labeled rollouts. \\[3pt]
\textbf{Cradle}\newline\citep{r2403_03186} & C, M, S & Joint\newline Sequential & On\newline Step, Event, Trajectory\newline Specialized & Environment, LLM feedback\newline Other & Instance result & A long-horizon computer-control agent that records observations and reflections, creates executable skills when reusable procedures emerge, and recalls them during later interaction. \\[3pt]
\textbf{DarwinX}\newline\citep{r2608_07545} & H, C, S & Teacher\newline Graph & Off\newline Batch\newline General, Specialized & Benchmark, Executable verifier, LLM feedback\newline Binary, Non-binary, LLM-as-a-judge, Other & Combined metrics & Branches and recombines harness versions, retaining complementary capabilities while using repeated evaluation and regression checks to control noisy improvements. \\[3pt]
\textbf{DCE + SRCL}\newline\citep{r2609_30652} & P & Joint\newline Sequential & Off\newline Batch\newline Specialized & Train/dev set, Executable verifier, LLM feedback\newline Binary, Ground truth, Other & No validation & Co-evolves a student and its privileged self-teacher, combining on-policy distillation with verified concise rewrites to improve subsequent reasoning rounds. \\[3pt]
\textbf{DemoEvolve}\newline\citep{r2605_24539} & H, C & Teacher\newline Graph & Off\newline Batch\newline Specialized & Train/dev set, Environment, Human\newline Non-binary, Other & Benchmark score & Uses competent human trajectories as references so a coding proposer can diagnose sparse, noisy failures and produce more stable harness edits in long-horizon environments. \\[3pt]
\textbf{Designer-RSI}\newline\citep{r2609_22086} & C, M, S & Teacher\newline Sequential & Off\newline Batch\newline Specialized & Train/dev set, Environment, LLM feedback\newline Binary, Non-binary, LLM-as-a-judge, Other & Benchmark score & Expands and repairs a graphic-design agent's procedural skill bank from user-task experience, admitting changes only when matched replays improve outcomes without regressing observed successes. \\[3pt]
\textbf{DGM}\newline\citep{r2505_22954} & H, C & Joint\newline Tree & Off\newline Batch\newline General & Benchmark, Executable verifier\newline Binary, Non-binary, Other & Combined metrics & Maintains an archive of coding-agent versions, samples parents, rewrites their harness code, and keeps empirically useful descendants as stepping stones for later evolution. \\[3pt]
\textbf{DIVE}\newline\citep{r2608_12486} & C, S & Self\newline Graph & Off\newline Batch\newline Specialized & Train/dev set, Executable verifier\newline Binary, Non-binary, Other & Benchmark score & Evolves diverse skill populations with a frozen model, combines skills across parents, and adapts the operators that generate new skills using measured utility. \\[3pt]
\textbf{DMoA}\newline\citep{dmoa} & O & Self\newline Sequential & Off, On\newline Batch\newline Specialized & Train/dev set, LLM feedback\newline Non-binary & No validation & Trains auxiliary routing weights from agent-output entropy so later queries use learned agent selection; the underlying language models remain frozen. \\[3pt]
\textbf{Dream-RSI}\newline\citep{dreamrsi} & H, C, M & Teacher\newline Sequential & On\newline Batch\newline Specialized & Benchmark, Environment, Executable verifier\newline Binary, Non-binary, Other & Combined metrics & Replays recorded discovery trees to evaluate and revise exploration-policy code cheaply, then redeploys the selected policy to guide fresh searches and expand the history available for improvement. \\[3pt]
\textbf{Dynamic Cheatsheet}\newline\citep{r2504_07952} & C, M, S & Teacher\newline Sequential & On\newline Trajectory\newline General, Specialized & LLM feedback\newline Other & No validation & Maintains a self-curated cheatsheet of strategies and code snippets across queries, updating memory from past attempts so later problems can reuse the accumulated experience. \\[3pt]
\textbf{Ecdysis}\newline\citep{r2609_11677} & H, C & Teacher\newline Sequential & Off\newline Batch\newline General & Train/dev set, Environment, Executable verifier, LLM feedback\newline Binary, Non-binary, LLM-as-a-judge, Other & Benchmark score & Turns batches of agent failures into repair specifications, edits executable runtime harnesses, and retains patches only when training-set performance improves. \\[3pt]
\textbf{EnvHarness}\newline\citep{r2608_19880} & C, M, S, O & Teacher\newline Sequential & Off\newline Batch\newline Specialized & Train/dev set, Environment, Executable verifier, LLM feedback\newline Binary, Non-binary, Other & Artifact validation, Combined metrics, No validation & Revises environment components from rollouts, checks solvability and behavioral measures before retention, and stores extracted skills for subsequent interaction and environment revision. \\[3pt]
\textbf{Escher-Loop}\newline\citep{r2604_23472} & H & Teacher\newline Graph & Off\newline Batch\newline General & Benchmark\newline Non-binary & Benchmark score & Maintains populations of task agents and optimizer agents; optimizers revise task programs and themselves while task-agent scores provide the signal for optimizer evolution. \\[3pt]
\textbf{Evo-Harness}\newline\citep{r2608_15071} & C, S & Joint\newline Sequential & On\newline Trajectory\newline Specialized & Train/dev set, LLM feedback\newline Binary, Other & Instance result & After each one-shot task, a solver exposes candidate experience and a separate evolver compiles useful context into structured harness skills for later tasks. \\[3pt]
\textbf{EvolveNet}\newline\citep{r2608_04968} & H, C & Teacher\newline Graph & Off\newline Batch\newline General, Specialized & Train/dev set, Executable verifier, LLM feedback\newline Binary, Non-binary, Other & Benchmark score & Evolves a shared harness on separate local workloads, combines evidence-backed changes with explicit scopes, and checks regressions before broadcasting the next version. \\[3pt]
\textbf{EvolveTrade}\newline\citep{r2609_17632} & C, S & Teacher\newline Sequential & On\newline Batch\newline Specialized & Environment, LLM feedback\newline Non-binary, Other & No validation & Periodically rewrites a trading agent's textual policy using its decisions, tool calls, rationales, and realized returns, carrying the revised instructions into later trading periods. \\[3pt]
\textbf{EvoOntology}\newline\citep{r2609_15779} & H, C, M & Teacher\newline Sequential & Off\newline Batch\newline Specialized & Train/dev set, Environment, Executable verifier, LLM feedback\newline Binary, Non-binary, Other & Benchmark score & Evolves a data agent's stored content, ontology schema, and executable tool layer from grounded probes and task traces, retaining changes through paired validation. \\[3pt]
\textbf{EvoSkill-GUI}\newline\citep{r2609_17653} & C, M, S & Self\newline Sequential & On\newline Event, Trajectory\newline Specialized & Environment, LLM feedback\newline Binary, LLM-as-a-judge, Other & Instance result & Revises reusable GUI skill packages during execution and after trajectory-level critique, storing improved procedures and recovery experience for similar later tasks. \\[3pt]
\textbf{EvoTrainer}\newline\citep{evotrainer} & P, H, M, S, O & Teacher\newline Graph & Off\newline Batch\newline General, Specialized & Train/dev set, Benchmark, Environment, Executable verifier, LLM feedback, Human\newline Binary, Non-binary, Other & Combined metrics & Uses rollouts to revise training recipes and diagnostic code, stores reusable lessons and skills, and promotes policy-training versions through performance and behavioral checks with human confirmation. \\[3pt]
\textbf{Experience Funnel}\newline\citep{r2609_08919} & P, C, M, S & Teacher\newline Sequential & Off\newline Batch\newline Specialized & Train/dev set, Environment, LLM feedback\newline Binary, Non-binary, Other & Benchmark score & Alternates validation-gated updates to a textual experience state with policy distillation, carrying both the improved state and student weights into the next round. \\[3pt]
\textbf{ExpGraph}\newline\citep{r2605_30712} & P, C, M, S & Teacher\newline Sequential & Off\newline Trajectory\newline General, Specialized & Train/dev set, Environment, Executable verifier\newline Binary, Non-binary, Other & Instance result & Converts task histories into a maintained experience graph and learns retrieval from measured gains over no-memory runs, while leaving the task executor frozen. \\[3pt]
\textbf{ForeDreamer}\newline\citep{r2608_20920} & H, C, M, S & Teacher\newline Graph & Off\newline Batch\newline Specialized & Train/dev set, Environment, Executable verifier\newline Binary, Non-binary, Other & Artifact validation, Benchmark score & Evolves forecasting lessons and evidence-processing instructions and tools, validating updates and reusing existing tools across procedural branches. \\[3pt]
\textbf{G-Zero}\newline\citep{r2605_09959} & P & Joint\newline Sequential & Off\newline Batch\newline General & LLM feedback\newline Non-binary & No validation & Co-evolves a Proposer and a Generator for open-ended generation: the Proposer is trained with GRPO to find queries and hints on which the Generator's hint-conditioned response shifts most, and the Generator is trained with DPO to internalize the hints, with no external judge. \\[3pt]
\textbf{GEA}\newline\citep{r2602_04837} & H, C & Joint\newline Graph & Off\newline Batch\newline General & Benchmark, Executable verifier, LLM feedback\newline Binary, Non-binary, Other & Artifact validation, Combined metrics & Pools code changes, execution traces, and outcomes across a group of archived agents, then uses the shared evidence to evolve each member’s harness. \\[3pt]
\textbf{GEPA}\newline\citep{r2507_19457} & C & Teacher\newline Graph & Off\newline Batch\newline Specialized & Train/dev set, Executable verifier, LLM feedback\newline Binary, Non-binary, Ground truth, Other & Benchmark score & Reflects on execution traces and task feedback to revise prompts, then evaluates and recombines complementary candidates through Pareto-based search. \\[3pt]
\textbf{Growing Harness}\newline\citep{r2609_26760} & H, C & Teacher\newline Sequential & Off\newline Batch\newline Specialized & Train/dev set, Environment, Executable verifier\newline Binary, Non-binary, Other & Artifact validation, Benchmark score & Grows a shared executable harness from failure-localized code edits, jointly repairs task windows, and rolls back changes that harm held-out success. \\[3pt]
\textbf{Gödel Agent}\newline\citep{r2410_04444} & H, C & Self\newline Sequential & Off\newline Batch\newline General & Benchmark, LLM feedback\newline Non-binary, Other & Benchmark score & Lets an agent inspect and rewrite its own executable logic, including the logic used for later self-modification, and empirically selects better descendants. \\[3pt]
\textbf{Harness-of-Harness}\newline\citep{r2609_01481} & C, M, O & Joint\newline Sequential & On\newline Trajectory\newline Specialized & Environment, Executable verifier, LLM feedback\newline Binary, LLM-as-a-judge, Other & Artifact validation & Runs repeated planning, development, and QA cycles over multiple days, preserving project state and verification evidence so later iterations can improve the same software project. \\[3pt]
\textbf{Harness-Zero}\newline\citep{r2609_24974} & H, C, M, S & Teacher\newline Sequential & Off\newline Batch\newline Specialized & Train/dev set, Benchmark, Environment, Executable verifier, LLM feedback\newline Binary, Non-binary, LLM-as-a-judge, Other & Artifact validation, Instance result & Revises harness tools, middleware, skills, and memory from recurring failures, then adapts the evolved harness into reference resources that guide student execution. \\[3pt]
\textbf{HarnessBank}\newline\citep{r2607_13683} & H, C & Teacher\newline Graph & Off\newline Batch\newline General, Specialized & Train/dev set, Environment, Executable verifier, LLM feedback\newline Binary, Non-binary, LLM-as-a-judge, Other & Artifact validation, Combined metrics & Maintains a semantically diverse bank of high-performing harnesses, reinvents or recombines their mechanisms, and admits offspring only after gated verification. \\[3pt]
\textbf{HarnessEvolve}\newline\citep{r2609_00829} & H, C, S & Teacher\newline Tree & Off\newline Batch\newline Specialized & Train/dev set, Environment, LLM feedback\newline Binary, Non-binary, Ground truth, LLM-as-a-judge, Other & Artifact validation, Combined metrics & Compares failed runs with verified reference trajectories to localize root causes, then accepts harness edits only after quality and recent-performance gates. \\[3pt]
\textbf{HarnessFix}\newline\citep{r2606_06324} & H, C, M & Teacher\newline Sequential & Off\newline Batch\newline Specialized & Benchmark, Executable verifier, LLM feedback\newline Binary, Non-binary, LLM-as-a-judge, Other & Artifact validation, Benchmark score & Compiles failed trajectories into a harness-aware representation, attributes failures to responsible steps and components, and applies scoped patches under regression-aware validation. \\[3pt]
\textbf{HarnessPAI}\newline\citep{r2609_29166} & H, C, M, S & Teacher\newline Sequential & Off\newline Batch\newline Specialized & Train/dev set, Environment, Executable verifier, LLM feedback\newline Binary, Non-binary, LLM-as-a-judge, Other & Artifact validation, Benchmark score & Revises robot execution programs from rollout videos and diagnostics, validates repairs across development scenes, and stores successful repairs as reusable skills. \\[3pt]
\textbf{HarnessX}\newline\citep{r2606_14249} & P, H, C, M & Teacher\newline Sequential & Off\newline Batch\newline General, Specialized & Benchmark, Environment, Executable verifier\newline Binary, Ground truth, Other & Artifact validation, Benchmark score & Composes typed harness primitives, evolves prompts, tools, memory, and control flow from execution traces, and can turn the resulting trajectories into model-training signals. \\[3pt]
\textbf{HeLa-Mem}\newline\citep{r2604_16839} & M & Teacher\newline Sequential & On\newline Event, Trajectory\newline General & Environment, LLM feedback\newline Other & No validation & Uses Hebbian co-activation to evolve an episodic memory graph and periodically distills dense memory hubs into reusable semantic knowledge. \\[3pt]
\textbf{HGM}\newline\citep{hgm} & H, C & Self\newline Tree & Off\newline Batch\newline General & Benchmark, Executable verifier, LLM feedback\newline Binary, Non-binary, Other & Benchmark score & Uses descendant performance to estimate a coding agent's self-improvement potential, guiding which archived agents rewrite themselves and which receive further evaluation. \\[3pt]
\textbf{Hyperagents}\newline\citep{r2603_19461} & H, C, M & Teacher\newline Graph & Off\newline Batch\newline General, Specialized & Benchmark, Executable verifier, LLM feedback\newline Binary, Non-binary, Other & Artifact validation, Benchmark score & Makes both the task agent and its meta-agent editable, allowing archived agents to improve task-solving code and the procedure used to generate later modifications. \\[3pt]
\textbf{HyperSkill}\newline\citep{r2608_16114} & C, M, S & Teacher\newline Sequential & On\newline Batch\newline Specialized & Environment, LLM feedback\newline Non-binary, Other & Instance result & Represents subtasks and reusable skills in a trajectory hypergraph, retrieves along two structural paths, and periodically prunes or merges skills using observed utility. \\[3pt]
\textbf{ISM}\newline\citep{r2606_31191} & M, S & Teacher\newline Sequential & On\newline Trajectory, Batch\newline Specialized & Benchmark, Executable verifier, LLM feedback\newline Binary, Other & Artifact validation, Instance result & Maintains a verified bank of mathematical strategies, using successes and failures to revise, merge, reinforce, and prune memories across isolated task episodes. \\[3pt]
\textbf{JAZ}\newline\citep{r2609_26891} & H, C, M, S & Teacher\newline Sequential & On\newline Trajectory\newline Specialized & Benchmark, Environment, Executable verifier\newline Binary, Non-binary, Other & No validation & Lets a meta-agent revise a solver's prompts and executable skills from task feedback, while a programmable context interface also supports long-horizon memory. \\[3pt]
\textbf{KSI}\newline\citep{r2607_19592} & C, M & Joint\newline Graph & On, Off\newline Trajectory, Batch\newline General, Specialized & Benchmark, Train/dev set, Environment, Executable verifier, LLM feedback\newline Binary, Ground truth, Other & Artifact validation & Keeps task agents disposable while evolving a shared knowledge base through evidence-grounded task forums, cross-task discussion, and distillation into guidance for later agents. \\[3pt]
\textbf{MCE}\newline\citep{r2601_21557} & H, C, M, S & Teacher\newline Graph, Sequential & Off, On\newline Batch, Trajectory\newline Specialized & Train/dev set, Benchmark, LLM feedback\newline Binary, Non-binary, Ground truth, Other & Artifact validation, Benchmark score & A meta-agent combines historical context-engineering skills, while a base agent applies them to revise reusable context files and retrieval code from task feedback; validation selects the best context. \\[3pt]
\textbf{MediSkill-Evo}\newline\citep{r2608_23397} & C, M, S & Teacher\newline Sequential & Off\newline Event, Trajectory\newline Specialized & Train/dev set, Environment, Executable verifier\newline Other & Artifact validation, Instance result & Writes clinical interaction experience into four typed knowledge banks only after provenance, scope, and process checks, then constrains later actions with the validated knowledge. \\[3pt]
\textbf{Mem$^2$Evolve}\newline\citep{r2604_10923} & H, C, M, S & Teacher\newline Sequential & On, Off$\to$On\newline Trajectory, Batch\newline General & Benchmark, LLM feedback\newline Non-binary, Ground truth, LLM-as-a-judge & Instance result & Co-evolves an experience memory and an asset memory: prior lessons guide the creation of tools or expert agents, whose use produces new experience for the next round. \\[3pt]
\textbf{Memp}\newline\citep{r2508_06433} & C, M, S & Teacher\newline Sequential & Off$\to$On\newline Trajectory\newline Specialized & Train/dev set, Environment\newline Other & Instance result & Distills trajectories into step-level instructions and script-like procedures, then continuously adds, corrects, and deprecates entries as experience accumulates. \\[3pt]
\textbf{MemRL}\newline\citep{memrl} & C, M, S & Teacher\newline Sequential & On, Off\newline Trajectory\newline Specialized & Train/dev set, Benchmark, Environment, Executable verifier, LLM feedback\newline Binary, Non-binary, Other & No validation & Stores trajectory-derived experiences with utility values, updates those values from task rewards, and combines utility with semantic relevance when retrieving guidance for later tasks. \\[3pt]
\textbf{Mendel Gödel Machine}\newline\citep{r2608_07645} & H, C & Joint\newline Graph & Off\newline Batch\newline General & Benchmark, Executable verifier, LLM feedback\newline Binary, Non-binary, Other & Combined metrics & Improves coding-agent descendants using evidence from several tasks or from another lineage, adding reaction-norm mutation and cross-lineage hybridization to archive search. \\[3pt]
\textbf{Meta-Harness}\newline\citep{r2603_28052} & H, C, M & Teacher\newline Graph & Off\newline Batch\newline Specialized & Benchmark, Executable verifier\newline Binary, Non-binary, Other & Combined metrics & Lets an agentic proposer inspect the code, scores, and execution traces of all prior candidates through a filesystem, then search for better task-specific harness code. \\[3pt]
\textbf{MM-Zero}\newline\citep{r2603_09206} & P & Joint\newline Sequential & Off\newline Batch\newline General & Executable verifier, LLM feedback\newline Binary, Non-binary & No validation & Trains Proposer, Coder, and Solver roles from one base VLM with GRPO: the Proposer poses visual concepts and questions, the Coder renders them into images as executable code, and the Solver reasons over them, so multimodal training data is generated from zero seed images. \\[3pt]
\textbf{ModularRSI}\newline\citep{r2609_14857} & H, C, M & Self\newline Graph & Off\newline Batch\newline General & Train/dev set, Environment, Executable verifier, LLM feedback\newline Binary, Other & Artifact validation & Separately evolves five harness modules from execution experience, checks candidate patches for validity, and integrates the retained modules into a general-purpose agent. \\[3pt]
\textbf{MoMHa}\newline\citep{r2609_30967} & H, C & Teacher\newline Graph & Off\newline Batch\newline Specialized & Train/dev set, Executable verifier, LLM feedback\newline Binary, Non-binary, Other & Artifact validation, Combined metrics & Uses historical harness code and execution traces to propose new harnesses, retaining candidates that improve a joint accuracy, safety, and token-cost objective. \\[3pt]
\textbf{MOSS}\newline\citep{r2605_22794} & H & Teacher\newline Sequential & On\newline Batch\newline General, Specialized & Environment, Executable verifier, LLM feedback, Human\newline Non-binary, LLM-as-a-judge, Other & Artifact validation, Combined metrics & Turns deployment failures into source-code revisions, tests rebuilt agents on captured tasks, and promotes fixes through consent and health-check gates. \\[3pt]
\textbf{NavHarness}\newline\citep{r2609_34276} & C, M, S & Joint\newline Sequential & On\newline Step, Event, Trajectory, Batch\newline Specialized & Environment, LLM feedback\newline Non-binary, LLM-as-a-judge, Other & No validation & Carries navigation experience across fresh sessions through revisable spatial memory, task handovers, and consolidated house notes and skills, using observations and model-based completion checks. \\[3pt]
\textbf{NeoHorse-1}\newline\citep{r2609_08183} & P & Teacher\newline Sequential & Off\newline Batch\newline General & Train/dev set, Environment, LLM feedback\newline Non-binary, Ground truth, Other & No validation & Uses routing and verification signals to organize agent experience for post-training, then reallocates the next training mixture according to the updated model's capability gaps. \\[3pt]
\textbf{OpenSkill}\newline\citep{r2606_06741} & H, C, S & Teacher\newline Sequential & Off\newline Batch\newline Specialized & Environment, Executable verifier, LLM feedback\newline Binary, Non-binary, Other & No validation & Builds skills and virtual tests from open-world resources, then uses execution diagnostics and targeted retrieval to refine skills before target-task use. \\[3pt]
\textbf{P$^2$O}\newline\citep{r2603_21877} & P, C & Self\newline Sequential & Off\newline Batch\newline General, Specialized & Benchmark, Executable verifier, LLM feedback\newline Binary, Ground truth, Other & Benchmark score & A self-improvement loop that alternates policy and prompt optimization: evolved prompts give policy learning better exploration templates, while the improving policy supplies increasingly difficult, high-value seeds for the next prompt round; prompt-induced gains are distilled into the model parameters. \\[3pt]
\textbf{PANDO}\newline\citep{r2605_24785} & H, C, M, S & Teacher\newline Sequential & On\newline Trajectory\newline Specialized & Environment, LLM feedback\newline Binary, Other & Instance result & Distills completed web interactions into reusable rules and executable routines, then updates confidence, merges redundant skills, and demotes unreliable ones during continued use. \\[3pt]
\textbf{PILOT}\newline\citep{r2608_26530} & C, M, S & Teacher\newline Graph & On\newline Event, Batch\newline General, Specialized & Environment, LLM feedback, Benchmark\newline Binary, Other & No validation, Instance result & A live supervisor steers active workers and updates reusable skills and memory during execution; successful runs contribute their updates to the shared state used in later iterations. \\[3pt]
\textbf{Prime Agent}\newline\citep{r2608_23552} & C, M, S & Self\newline Sequential & On\newline Step, Event, Trajectory\newline General, Specialized & Environment, Executable verifier\newline Other & Instance result & A persistent RLM-style harness that preserves histories, memories, skills, prompts, and subagent specifications across trajectories so later work can build on earlier context. \\[3pt]
\textbf{Procedural Graphs}\newline\citep{r2609_09153} & C, M, S & Teacher\newline Sequential & Off\newline Batch\newline Specialized & Train/dev set, Environment, LLM feedback\newline Binary, Non-binary, Other & Artifact validation, Benchmark score & Builds a procedural graph from successful and failed trajectories, validates graph edits before retaining them, and turns relevant subgraphs into step-level guidance for later tasks. \\[3pt]
\textbf{R-Few}\newline\citep{r2512_02472} & P & Joint\newline Sequential & Off\newline Batch\newline General & Train/dev set, LLM feedback, Human\newline Binary, Non-binary & No validation & Guides Challenger–Solver self-play with lightweight human oversight: the Challenger samples a few human-labeled examples to ground question generation, and the Solver trains on mixed human and synthetic data under an adaptive difficulty-based curriculum, which mitigates drift and diversity collapse. \\[3pt]
\textbf{R-Zero}\newline\citep{r2508_05004} & P & Joint\newline Sequential & Off\newline Batch\newline General & LLM feedback\newline Binary & No validation & Initializes a Challenger and a Solver from one base model; the Challenger is rewarded for questions at the edge of the Solver's ability and the Solver trains on majority-vote pseudo-labels, so the curriculum tracks the Solver across rounds without any external data. \\[3pt]
\textbf{ReasoningBank}\newline\citep{r2509_25140} & C, M & Joint\newline Sequential & On\newline Trajectory\newline General, Specialized & Environment, LLM feedback\newline Binary, Other & No validation & Distills generalizable reasoning strategies from self-judged successes and failures, retrieves them for later tasks, and uses additional test-time rollouts to improve the memory bank. \\[3pt]
\textbf{Recuris}\newline\citep{r2608_24876} & C, M, S & Teacher\newline Sequential & Off, Off$\to$On\newline Trajectory\newline Specialized & Environment, Executable verifier, LLM feedback\newline Other & Artifact validation, Instance result & Couples working and experiential memory so long-horizon execution localizes memory failures, then lets a fixed meta-agent make validation-gated updates to skill memory. \\[3pt]
\textbf{Red Queen Gödel Machine}\newline\citep{r2606_26294} & H, C & Teacher\newline Tree & Off\newline Batch\newline General & Benchmark, LLM feedback\newline Non-binary, LLM-as-a-judge & Combined metrics & Extends archive-based self-improvement so the agent and its utility or evaluator can both evolve, while preserving stable selection within each evaluation epoch. \\[3pt]
\textbf{RHI}\newline\citep{r2607_15524} & C & Teacher\newline Sequential & Off\newline Batch\newline Specialized & Benchmark, LLM feedback\newline Non-binary, LLM-as-a-judge & Benchmark score & Represents an agent loop as a prompt-level harness and repeatedly revises it using pairwise feedback accumulated across consecutive versions. \\[3pt]
\textbf{RISE}\newline\citep{rise} & P & Self\newline Graph & Off\newline Batch\newline General, Specialized & Train/dev set, Environment, Executable verifier, LLM feedback\newline Binary, Non-binary, Other & No validation & Alternates verifiable-reward training with distillation from a teacher extrapolated from its own checkpoints, refreshing the teacher as the student improves. \\[3pt]
\textbf{RRSI}\newline\citep{rrsi} & H, C, M, S & Teacher\newline Graph & Off\newline Batch\newline General & Train/dev set, Environment, Executable verifier, LLM feedback\newline Binary, Non-binary, LLM-as-a-judge, Other & Combined metrics & Regularizes harness evolution with shrinking edit budgets, history-guided proposals, leakage checks, and cost-aware selection to reduce overfitting to evolution tasks. \\[3pt]
\textbf{RSIAgent}\newline\citep{r2609_15364} & C, M, S & Self\newline Sequential & Off\newline Trajectory, Batch\newline Specialized & Environment, LLM feedback\newline Binary, LLM-as-a-judge, Other & Instance result & Explores unfamiliar environments with curriculum, actor, and verifier roles, updates reusable memory from verified experience, and freezes that memory for downstream execution. \\[3pt]
\textbf{SAGEAgent}\newline\citep{r2607_09521} & C, M, S & Teacher\newline Sequential & Off\newline Trajectory, Batch\newline Specialized & Train/dev set, Environment, LLM feedback\newline Non-binary, Other & Artifact validation & Accumulates decision episodes and periodically revises explicit acquisition rules using cost and uncertainty feedback, improving which diagnostic modalities a frozen agent requests. \\[3pt]
\textbf{SEAL}\newline\citep{r2506_10943} & P & Self\newline Sequential & Off\newline Batch\newline General & Train/dev set\newline Non-binary, Ground truth & Benchmark score & The model writes self-edits that specify its own finetuning data and update directives; downstream performance rewards self-edits that produce useful weight changes. \\[3pt]
\textbf{SEEK}\newline\citep{r2609_29803} & C, S & Teacher\newline Sequential & On\newline Batch\newline Specialized & Train/dev set, Human, LLM feedback\newline Non-binary, Ground truth, Other & Benchmark score & Uses reviewed search-evaluation errors to revise a skill bank, accepting changes only when they improve new cases without degrading historical replay performance. \\[3pt]
\textbf{Self-Harness}\newline\citep{r2606_09498} & H, C, M & Self\newline Graph & Off\newline Batch\newline Specialized & Benchmark, Executable verifier, LLM feedback\newline Binary, Non-binary, Other & Combined metrics & Uses the same fixed model to mine recurring failures, propose bounded edits to its own harness, and merge only candidates that pass held-in and held-out regression tests. \\[3pt]
\textbf{SePO}\newline\citep{r2606_04465} & C & Self, Teacher\newline Tree & Off\newline Batch\newline General, Specialized & Train/dev set, Executable verifier\newline Binary, Non-binary & Benchmark score & Evolves a prompt agent's own system prompt on a multi-task pool, then uses that improved prompt optimizer to evolve task-agent prompts for target tasks. \\[3pt]
\textbf{SIFT}\newline\citep{sift} & H, C & Teacher\newline Tree & Off\newline Batch\newline General & Benchmark, Executable verifier, LLM feedback\newline Binary, Non-binary, Other & Combined metrics & Searches a tree of coding-agent versions using pairwise code judgments alongside task evaluations, reducing the number of expensive benchmark runs needed to choose promising branches. \\[3pt]
\textbf{SimSkill}\newline\citep{r2609_03753} & H, C, M, S & Teacher\newline Sequential & Off, On\newline Trajectory\newline Specialized & Environment, Executable verifier, LLM feedback\newline Binary, LLM-as-a-judge, Other & Artifact validation, Instance result & Explores traffic-simulation tasks through an action–critic loop and consolidates completed episodes into reusable code skills and linked knowledge pages. \\[3pt]
\textbf{SkillAA}\newline\citep{r2609_20455} & C, M, S & Teacher\newline Sequential & Off\newline Batch\newline Specialized & Train/dev set, Executable verifier, LLM feedback\newline Binary, Non-binary, Ground truth, Other & Benchmark score & Attributes task failures to nodes in a skill graph, proposes local edits, and applies targeted and full-pool validation with rollback before retaining an update. \\[3pt]
\textbf{SkillAdam}\newline\citep{r2609_08944} & C, M, S & Teacher\newline Sequential & Off\newline Batch\newline Specialized & Train/dev set, Environment, Executable verifier, LLM feedback\newline Binary, Non-binary, LLM-as-a-judge, Other & Combined metrics & Revises skill documents using a persistent issue tracker and an adaptive edit budget, retaining changes when batch-level evaluation shows gains without unacceptable regressions. \\[3pt]
\textbf{SkillEvoReg}\newline\citep{r2609_30861} & H, C, S & Teacher\newline Sequential & Off, On\newline Trajectory, Batch\newline Specialized & Train/dev set, Benchmark, Executable verifier, LLM feedback\newline Binary, Non-binary, Other & Artifact validation, Instance result, Benchmark score & Regularizes repeated skill updates using skill dropout, localized complexity control, and paired counterexample checks to reduce overfitting and regressions. \\[3pt]
\textbf{SkillGLoW}\newline\citep{r2609_02217} & C, M, S & Teacher\newline Graph & On, Off\newline Trajectory, Batch\newline Specialized & Benchmark, Train/dev set, Environment, Executable verifier, LLM feedback\newline Binary, Non-binary, Other & Benchmark score & Groups task-local experience by shared solving procedure, consolidates it into reusable skill priors, and checks library revisions through execution before combining them with fresh local skills. \\[3pt]
\textbf{SkillHEX}\newline\citep{r2608_05628} & C, S & Teacher\newline Graph & On\newline Trajectory\newline Specialized & Benchmark, Environment, Executable verifier, LLM feedback\newline Binary, Non-binary, Other & Artifact validation, Instance result & Tests explicit failure hypotheses against cached outputs and shares the resulting evidence across skill-revision branches to guide budget-limited test-time improvement. \\[3pt]
\textbf{SkillPivot}\newline\citep{r2609_29154} & C, S & Teacher\newline Sequential & On\newline Batch\newline Specialized & Benchmark, Environment, LLM feedback\newline Binary, Non-binary, LLM-as-a-judge, Other & Combined metrics & Locates where a failed trajectory went off track, compares it with a teacher's continuation from the same prefix, and retains localized skill edits after preservation and generalizability checks. \\[3pt]
\textbf{SkillProx}\newline\citep{skillprox} & C, M, S & Teacher\newline Graph & Off\newline Batch\newline Specialized & Train/dev set, Environment, Executable verifier, LLM feedback\newline Binary, Non-binary, LLM-as-a-judge, Other & Artifact validation, Combined metrics & Revises skills from execution feedback, then removes or consolidates low-utility content through edits checked for structural validity, lower complexity, and preserved task performance. \\[3pt]
\textbf{SkillRefine}\newline\citep{r2609_30674} & C, M, S & Teacher\newline Sequential & Off\newline Batch\newline Specialized & Train/dev set, Environment, Executable verifier, LLM feedback\newline Binary, Non-binary, Ground truth, Other & No validation & Builds refinery-planning skills from documentation and expert records, then uses execution constraints and reference comparisons to repair skill content and retrieval metadata. \\[3pt]
\textbf{SkillRevise}\newline\citep{r2606_01139} & C, S & Teacher\newline Sequential & On\newline Trajectory\newline Specialized & Benchmark, Environment, Executable verifier, LLM feedback\newline Binary, Non-binary, LLM-as-a-judge, Other & Combined metrics & Diagnoses skill defects from execution traces and verifier feedback, revises the skill with repair principles, and returns the best tested candidate within a bounded budget. \\[3pt]
\textbf{SkillSmith}\newline\citep{r2606_01314} & H, M, S & Teacher\newline Graph & Off\newline Batch\newline General & Train/dev set, Executable verifier, LLM feedback\newline Non-binary, Other & Artifact validation, Combined metrics & Reflects on failures to jointly edit skills and executable tools, validates bundled changes progressively, and records anti-patterns to prevent repeated mistakes. \\[3pt]
\textbf{SoL-Pi}\newline\citep{r2609_20519} & H, C & Teacher\newline Graph & Off\newline Batch\newline General & Train/dev set, Environment, Executable verifier, LLM feedback\newline Binary, Non-binary, LLM-as-a-judge, Other & Combined metrics & Scales parallel research loops over harness changes, validates efficiency gains under capability constraints, and combines retained mechanisms into a lower-cost agent harness. \\[3pt]
\textbf{TextGrad}\newline\citep{textgrad} & C & Teacher\newline Sequential & Off\newline Batch\newline Specialized & Train/dev set, Executable verifier, LLM feedback\newline Binary, Non-binary, Ground truth, LLM-as-a-judge & Benchmark score & Backpropagates textual feedback through a computation graph to revise reusable prompts, retaining updates that improve validation performance. \\[3pt]
\textbf{TISD}\newline\citep{r2609_30878} & P & Joint\newline Sequential & Off\newline Batch\newline Specialized & Train/dev set, Executable verifier, LLM feedback\newline Binary, Other & No validation & Uses teacher–student disagreement to branch a student rollout, regenerates its continuation, and distills the resulting trajectories while updating the teacher from the student. \\[3pt]
\textbf{TRACE}\newline\citep{r2608_22793} & C, S & Teacher\newline Sequential & Off\newline Batch\newline Specialized & Benchmark, LLM feedback\newline Binary, Other & Benchmark score & Groups trajectories by invoked skills and contrasts successful with failed behavior after each development evaluation round to rewrite a reusable behavioral skill bank. \\[3pt]
\textbf{TTCS}\newline\citep{r2601_22628} & P & Joint\newline Sequential & On\newline Batch\newline Specialized & LLM feedback\newline Binary & No validation & Co-evolves a question synthesizer and a solver during test-time training: the synthesizer generates progressively harder variants of the test questions guided by solver feedback, and the solver updates its weights with self-consistency rewards on both original and synthesized questions. \\[3pt]
\textbf{TTSE}\newline\citep{r2609_24289} & C, M, S & Teacher\newline Sequential & Off, On\newline Batch, Trajectory\newline Specialized & Train/dev set, Benchmark, Environment, LLM feedback\newline Binary, Non-binary, LLM-as-a-judge, Other & No validation, Instance result & Separates environmental facts from conditional procedures, then uses task feedback to retire misleading rules, resolve contradictions, and update both banks for later tasks. \\[3pt]
\textbf{VisPlay}\newline\citep{r2511_15661} & P & Joint\newline Sequential & Off\newline Batch\newline General & LLM feedback\newline Binary & No validation & Assigns one base VLM the roles of an image-conditioned Questioner and a multimodal Reasoner, jointly trained with GRPO using diversity and difficulty rewards, so the model improves visual reasoning from unlabeled images alone. \\[3pt]
\textbf{When Rules Learn}\newline\citep{r2606_17220} & C, S & Self\newline Sequential & Off\newline Batch\newline Specialized & Train/dev set, Executable verifier\newline Non-binary & Benchmark score & An agent creates query-rewriting rules, plans experiments over rule combinations, and removes rules that fail to improve legal retrieval metrics. \\[3pt]
\textbf{WikiSkill}\newline\citep{r2608_27454} & C, M, S & Teacher\newline Sequential & Off\newline Batch\newline Specialized & Train/dev set, Environment, LLM feedback\newline Binary, Non-binary, Ground truth, Other & Benchmark score & Maintains an immutable raw-trace layer and a persistent wiki that survives rejected skill edits, allowing later skill updates to build on accumulated evidence. \\[3pt]
\end{longtable}\endgroup

%% file: tables/benchmarks.tex
\begingroup\footnotesize\setlength{\tabcolsep}{3pt}\renewcommand{\arraystretch}{1.12}
\begin{longtable}{@{}>{\raggedright\arraybackslash}p{0.19\dimexpr\linewidth-10\tabcolsep\relax}>{\raggedright\arraybackslash}p{0.07\dimexpr\linewidth-10\tabcolsep\relax}>{\raggedright\arraybackslash}p{0.07\dimexpr\linewidth-10\tabcolsep\relax}>{\raggedright\arraybackslash}p{0.25\dimexpr\linewidth-10\tabcolsep\relax}>{\raggedright\arraybackslash}p{0.24\dimexpr\linewidth-10\tabcolsep\relax}>{\raggedright\arraybackslash}p{0.18\dimexpr\linewidth-10\tabcolsep\relax}@{}}
\caption{Benchmarks and their evaluation settings.}\label{tab:benchmarks}\\
\multicolumn{6}{@{}p{\linewidth}@{}}{P: parameters; H: harness code; C: context; M: memory; S: skills; O: other artifacts. Off/On: offline/online.}\\[3pt]
\toprule
Benchmark & Resources & Mode & Task setup & Feedback for updates & Reported measures\\\midrule\endfirsthead
\multicolumn{6}{@{}l}{\tablename\ \thetable\ (continued)}\\[3pt]
\multicolumn{6}{@{}p{\linewidth}@{}}{P: parameters; H: harness code; C: context; M: memory; S: skills; O: other artifacts. Off/On: offline/online.}\\[3pt]
\toprule Benchmark & Resources & Mode & Task setup & Feedback for updates & Reported measures\\\midrule\endhead
\midrule\multicolumn{6}{r@{}}{\textit{Continued on the next page.}}\\\endfoot
\bottomrule\endlastfoot
LifelongAgentBench \citep{r2505_11942} & C, M & On & Skill-dependent task sequences in interactive environments & Environment observations and terminal success/failure & Task performance, Gain, Cost \\[3pt]
StuLife \citep{r2508_19005} & P, C, S, M & On & Stateful campus task sequence; includes a fine-tuning baseline & Environment observations, tool results, and task outcomes & Task performance, Gain, Interaction turns \\[3pt]
Evo-Memory \citep{r2511_20857} & C, M & On & Task streams with varied order and difficulty & Task trajectories and correctness signals & Task performance, Gain, Interaction steps \\[3pt]
SkillFlow \citep{r2604_17308} & C, S & On & Task families ordered by within-family difficulty & Execution traces and rubric-based verifier feedback & Task performance, Gain, Cost \\[3pt]
CL-Bench \citep{r2606_05661} & C, M, O & On & Related instances in shared environments, including concept drift & Environment observations and task-dependent feedback & Task performance, Gain, Cost \\[3pt]
AgentStream \citep{r2608_00155} & C, S, M & On & Isolated, sequential, and interleaved domain streams & Execution outcomes and self-evaluation & Task performance, Gain, Cost \\[3pt]
PATH-Bench \citep{r2608_01149} & C, S, M & On & Repeated probes within helpful or interfering task histories & Intervening-task experience; probes do not update state & Task performance, Transfer, Forgetting \\[3pt]
PAST-Bench \citep{r2608_04003} & S, M & On & Learn, reuse, and control episodes in fresh sessions & Rules, procedures, and corrections in learning episodes & Task performance, Gain \\[3pt]
ContinualSkillBench \citep{r2608_03874} & C, S & On & Tasks ordered by difficulty and skill dependencies & Execution trajectories and evaluator feedback & Task performance, Gain \\[3pt]
FinEvo-Bench \citep{r2608_06144} & S, M & On & Shuffled, interleaved streams of related professional cases & Rubric-based feedback after task completion & Task performance, Gain, Cost \\[3pt]
PostTrainBench \citep{r2603_08640} & C, O & On & Research-agent loop over target-model post-training; final checkpoint test & Execution logs and agent-run experiments & Task performance, Gain, Cost, Time \\[3pt]
Agent$^2$ RL-Bench \citep{r2604_10547} & C, O & On & Research-agent loop with repeated training and target-model submissions & Grading-server scores for submitted models & Task performance, Gain, Cost, Time \\[3pt]
MLS-Bench \citep{r2605_08678} & C, O & On & Research-agent loop for ML-method components across evaluation settings & Training and visible-test metrics returned by experiment tools & Task performance, Gain, Cost \\[3pt]
Curation-Bench \citep{r2606_04261} & C, O & On & Research-agent loop over data-curation policies for a target model & Scores from fixed training and evaluation pipelines & Task performance, Gain, Cost \\[3pt]
AutoLab \citep{r2606_05080} & C, O & On & Iterative improvement of a working implementation within a time budget & Execution results and local benchmark measurements & Task performance, Gain, Cost, Time \\[3pt]
EdgeBench \citep{r2607_05155} & C, O & On & Repeated attempts within a long-running task & Local tests and submission-gated judge feedback & Task performance, Gain, Cost, Time \\[3pt]
RSIBench-Data \citep{r2607_25886} & C, O & On & Research-agent loop revising training-data strategies for a target model & Checkpoint scores, trajectories, and execution diagnostics & Task performance, Gain, Cost \\[3pt]
MemoryAgentBench \citep{r2507_05257} & C, M & Off & Incremental context ingestion, then question answering & Provided context chunks and labeled examples & Task performance, Gain, Cost, Time \\[3pt]
MemoryBench \citep{r2510_17281} & P, M & Off & Feedback training, then held-out tests; includes weight fine-tuning & Simulated user comments and actions & Task performance, Gain, Cost, Time \\[3pt]
VeRO-Bench \citep{r2602_22480} & H, C & Off & Optimize target harnesses, then evaluate held-out tasks & Development scores and structured execution traces & Task performance, Gain, Cost \\[3pt]
SkillLearnBench \citep{r2604_20087} & S & Off & Generate skills from seed tasks, then test held-out instances & Seed executions and self/teacher feedback, depending on method & Task performance, Gain, Cost \\[3pt]
Meta-Agent Challenge \citep{r2606_04455} & H, C & Off & Agent development followed by held-out verification & Development-set correctness and aggregate scores & Task performance, Cost, Time \\[3pt]
SEAGym \citep{r2606_17546} & H, C, M, S & Off & Training-task updates; frozen-snapshot validation and transfer tests & Training trajectories and visible verifier feedback & Task performance, Gain, Cost, Time \\[3pt]
EvoAgentBench \citep{r2607_05202} & S & Off & Training and held-out tasks linked by shared procedural abilities & Training trajectories and verifier outcomes & Task performance, Gain, Cost \\[3pt]
GDPevo \citep{r2608_03764} & S & Off & Training rules recombined in held-out workflows & Reference answers, grader scores, or environment observations & Task performance, Gain, Cost \\[3pt]
HarnessOpt-Bench \citep{r2608_06301} & H, C & Off & Select a revised harness, then test held-out examples & Development traces/results and aggregate validation scores & Task performance, Gain, Cost \\[3pt]
Evo-Bench \citep{r2608_09096} & H, C & Off & Harness-sensitive tasks split for optimization and final testing & Optimization scores, trajectories, and diagnostics & Task performance, Gain, Cost \\[3pt]
S3Gym \citep{r2608_31100} & P, C, M & Off & Game exploration followed by stricter held-out evaluation & Environment observations and self-judgments & Task performance, Gain \\[3pt]
HarnessDev \citep{r2609_01437} & H, C & Off & Harness creation/evolution before testing frozen code & Development cases and downstream execution feedback & Task performance, Gain, Cost \\[3pt]
EvoMemBench \citep{r2605_18421} & C, M & On, Off & Online within/across episodes; frozen-memory transfer across environments & Incoming information, tool results, and prior trajectories & Task performance, Gain, Cost \\[3pt]
Aspire \citep{r2608_31111} & P, H, C, M & On, Off & Final-only and adaptive-feedback protocols; weight or harness tracks & Agent-created validation data; aggregate scores in adaptive protocol & Task performance, Gain, Cost, Time \\[3pt]
\end{longtable}\endgroup